%% file: main.tex
\documentclass[preprint]{elsarticle}

\usepackage{listings}
\usepackage{placeins}
\usepackage{amsmath, amssymb}
\usepackage{color}
\usepackage{xcolor}
\usepackage{lineno}
\usepackage{makecell}
\usepackage{multirow}
\usepackage{hyperref}       
\usepackage{url}            
\usepackage{amsfonts}       
\usepackage{nicefrac}       
\usepackage{microtype}      
\usepackage{doi}
\usepackage{dcolumn}
\usepackage{bm}
\usepackage{caption}
\usepackage{subcaption}
\usepackage{float}
\usepackage{array}
\newcolumntype{P}[1]{>{\centering\arraybackslash}p{#1}}

\definecolor{lightblue}{RGB}{240,245,255}
\definecolor{darkblue}{RGB}{40,40,85}

\journal{}

\date{July 1, 2025} 

\begin{document}

\begin{frontmatter}

\title{JAX-MPM: A Learning-Augmented Differentiable Meshfree Framework for GPU-Accelerated Lagrangian Simulation and Geophysical Inverse Modeling}

\author[UMN]{Honghui Du}

\author[UMN]{QiZhi He\corref{mycorrespondingauthor}}
\cortext[mycorrespondingauthor]{Corresponding author}
\ead{qzhe@umn.edu}
\address[UMN]{Department of Civil, Environmental, and Geo- Engineering, University of Minnesota, 500 Pillsbury Drive S.E., Minneapolis, MN 55455}

\begin{abstract}
Differentiable programming has emerged as a powerful paradigm in scientific computing, enabling automatic differentiation through simulation pipelines and naturally supporting both forward and inverse modeling.
We present JAX-MPM,
a general-purpose differentiable meshfree solver based on the material point method (MPM) and implemented in the modern JAX architecture.
The solver adopts a hybrid Eulerian–Lagrangian framework 
to capture large deformations, frictional contact, and inelastic material behavior, with emphasis on geomechanics and geophysical hazard applications.
Leveraging GPU acceleration and automatic differentiation, JAX-MPM enables efficient gradient-based optimization directly through its time-stepping solvers and supports
joint training of physical models with deep learning  to infer unknown system conditions and uncover hidden constitutive parameters.
We validate JAX-MPM  through a series of 2D and 3D benchmark simulations, including dam-break and granular collapse problems, demonstrating both numerical accuracy and GPU-accelerated performance. 
Results show that a high-resolution 3D granular cylinder collapse with 2.7 million particles completes 1000 time steps in approximately 22 seconds (single precision) and 98 seconds (double precision) on a single GPU.
Beyond high-fidelity forward modeling, 
we demonstrate the framework’s inverse modeling capabilities through tasks such as velocity field reconstruction and the estimation of spatially varying friction from sparse data.
In particular, JAX-MPM accommodates data assimilation from both Lagrangian (particle-based) and Eulerian (region-based) observations, and can be seamlessly coupled with neural network representations.
These  results establish JAX-MPM as a unified and scalable differentiable meshfree platform, that advances fast physical simulation and data assimilation for complex solid and geophysical systems.

\end{abstract}


\end{frontmatter}

\input{Intro_new}
\input{Preliminaries}

\input{Approximation}
\input{results}

\section{Conclusion}\label{sec:conclusion}

This study presents JAX-MPM, a fully differentiable, GPU-accelerated material point method (MPM) framework tailored for large-deformation predictive simulation and inverse modeling in continuum mechanics, with a focus on geophysical flow applications. 
Built entirely on the JAX ecosystem, the framework leverages modern differentiable programming capabilities, including automatic differentiation, JIT compilation, and vectorized GPU execution, to support scalable, end-to-end optimization pipelines. 
JAX-MPM also integrates naturally with neural network libraries, enabling hybrid physical–ML workflows for data-driven modeling and field reconstruction.

We validated JAX-MPM on a suite of forward benchmarks, including 2D and 3D dam-break problems and granular collapses, demonstrating high accuracy, computational efficiency, and scalability. 
Compared to conventional CPU-based implementations in C++ or Fortran, JAX-MPM achieves substantial speedups (up to 100×) by exploiting GPU acceleration. 
Using JAX-MPM, we further investigated the influence of initial geometry (e.g., aspect ratio) and material properties (e.g., internal friction angle) on granular flow dynamics. The results show strong agreement with theoretical predictions and experimental observations, underscoring the effectiveness of the framework for realistic geophysical flow simulation and its potential to advance the understanding and design of processes.

Beyond forward simulation, JAX-MPM enables inverse modeling through a unified formulation that supports both particle-based (Lagrangian) and region-based (Eulerian) supervision. To evaluate this capability, we conducted several inverse analysis tasks, including: (1) reconstruction of initial velocity fields, both parametric and non-parametric (neural network-based), from sparse trajectory or velocity monitor data; (2) estimation of spatially varying friction coefficients along the base of a dam-break domain, even under contact-dominated dynamics. 
These experiments demonstrate that JAX-MPM can recover localized and spatially varying parameters from limited observations with high fidelity. 
To the best of our knowledge, this represents the first demonstration of a differentiable physics-based solver applied to large-scale Lagrangian geophysical flows for inverse frictional parameter estimation.
This makes it a versatile tool for real-world geomechanical applications, where direct measurement is often infeasible and observational data is sparse.

Furthermore, we showcase the integration of feed-forward neural networks within JAX-MPM to approximate unknown parameter fields (Section \ref{sec:inverse_velocity_nn}).
The successful reconstruction illustrates the potential of neural parameterizations for capturing complex functional relationships, benefiting from universal approximation and smooth differentiability. 
The framework’s general differentiability also opens opportunities for incorporating alternative modeling paradigms such as symbolic regression, Gaussian process regression, or hybrid symbolic–neural formulations (e.g., NIM~\cite{du2024neural}) when prior knowledge is available.

In summary, JAX-MPM advances the field of  differentiable physics for continuum mechanics by bridging high-fidelity simulation with scalable gradient-based optimization. It offers a flexible and efficient foundation for future research in inverse modeling, learning-enabled simulation, and digital twins for geohazard analysis, natural hazard mitigation, and intelligent design of geotechnical systems.

\input{appendix}

\bibliographystyle{cas-model2-names} 

\bibliography{ref_new}

\end{document}

%% file: Intro_new.tex
\section{Introduction}
\subsection{Overview}

Natural hazards, or geohazards \cite{gill2014reviewing,highland1997debris}, such as landslides, dam-break flows, debris flows, and soil liquefaction, 
are 
gravitationally driven processes involving large-scale failure of geomaterials (e.g., soils, rocks, and granular media in the Earth's near-surface) and 
they pose
significant risks to both natural and built environments.
Governed by highly nonlinear, path-dependent material behavior \cite{wood1990soil}, these phenomena often entail severe deformations and rapidly evolving boundary conditions~\cite{augarde2021numerical}, making high-fidelity predictive modeling and simulation particularly challenging.

Numerical modeling and simulation are essential tools in geomechanics and geotechnical engineering, enabling the analysis of complex physical processes through the solution of partial differential equations (PDEs). 
Mesh-based methods such as the finite element method (FEM) \cite{potts2001finite, lees2016geotechnical, wang2015large} and finite volume method (FVM) \cite{demirdvzic1995numerical, deb2017finite} have been widely adopted for this purpose. For example, FEM has been applied to simulate flow-like landslides \cite{crosta2003numerical}, while FVM has been used to model coupled flow and shear failure in fractured geomechanical systems \cite{deb2017finite}. These methods have provided valuable insights in geohazard research and remain effective for problems involving moderate deformation and well-defined boundaries.
However, in geohazard modeling scenarios characterized by extreme deformation, fragmentation, and evolving topologies, mesh-based methods often face numerical challenges due to severe mesh distortion and tangling, and can become unreliable or computationally impractical. To address these limitations, meshfree methods have been developed, representing materials as collections of particles or discrete points—eliminating the need for conforming meshes and improving robustness in large-deformation problems.

Another critical challenge in geomechanical modeling is the limited observability of real-world systems. For instance, essential quantities, such as initial velocity fields, internal stress states, and spatially varying friction, are usually unknown or unmeasurable in situ.
Therefore, inverse modeling becomes essential in these cases, enabling the estimation of latent physical parameters from partial observations~\cite{calvello2017role, abraham2021runout}, to improve predictive accuracy and support real-world risk assessment and design.
Conventional approaches typically rely on repeated forward simulations within optimization loops, including probabilistic methods such as Bayesian inference with Gaussian processes~\cite{zhao2022bayesian} and Markov chain Monte Carlo (MCMC) methods~\cite{chen2022inverse}. While these offer uncertainty quantification, they are computationally expensive and difficult to scale in high-dimensional, nonlinear systems. Gradient-based alternatives such as adjoint solvers improve efficiency, but are often impractical to implement for systems with contact, friction, and path-dependent materials.

Altogether, the above-mentioned challenges highlight the need for a fundamentally novel simulation framework that integrates high-fidelity physical models, capable of capturing complex deformation and material inelasticity,
with inverse modeling capability. 

\subsection{Lagrangian meshfree simulation}
Meshfree methods provide a robust foundation for handling large deformations and dynamic boundary conditions, 
as they naturally avoid mesh entanglement and eliminate the need for frequent remeshing~\cite{chen2017meshfree}. 
Prominent examples include smoothed particle hydrodynamics (SPH) \cite{gingold1977smoothed, liu2003smoothed}, the discrete element method (DEM) \cite{cundall1979discrete, cundall1992numerical}, and the material point method (MPM) \cite{sulsky1994particle, sulsky1995application, kumar2019scalable}. 
These methods commonly utilize a Lagrangian formulation,  in which computational particles move with the material, explicitly carrying essential state variables such as stress, strain, and  internal history variables. 
This particle-based representation inherently supports accurate tracking of material evolution and path-dependent behavior, which is crucial for capturing complex phenomena in geomechanical simulations.

Among these methods, MPM 
stands out due to its hybrid Eulerian–Lagrangian formulation. 
In this approach, material points carry internal state variables and track continuum deformation in a Lagrangian manner, while computational operations are performed on a background Eulerian grid that is reset at each time step. 
This decoupling of material and computational domains offers notable advantages~\cite{de2020material}, including avoiding mesh distortion 
and efficiently handling complex contact and  costly particle–particle interactions
via the background grid.
These attributes make MPM particularly attractive for modeling challenging geomechanical problems~\cite{bardenhagen2000material, andersen2010modelling, bandara2015coupling}
such as landslides, debris flows, and multiphase failure, 
where large deformations, fragmentation, and multi-body interactions are dominant. 
Given these strengths, 
MPM is adopted as the computational foundation for the differentiable simulation framework developed in this study.

\subsection{Machine learning and differentiable simulators in geomechanics}

To address the challenges of inverse modeling in geophysical hazards, where traditional methods often face limitations in efficiency and scalability,
we explore differentiable programming as a flexible and scalable alternative.
Differentiable programming (DP)~\cite{innes2019differentiable,rackauckas2020universal,blondel2024elements} enables end-to-end gradient computation through entire simulation pipelines using automatic differentiation (AD)~\cite{baydin2018automatic}.
Unlike sampling-based or adjoint-based approaches, DP avoids the need for manual derivations and naturally supports optimization over both physical parameters and latent states, even in systems with complex dynamics.

DP-based methods can be broadly categorized according to how differentiability is achieved—either through continuous formulations or discretized numerical solvers.
The first category leverages neural networks (NNs) to achieve differentiability, for example, by embedding physical laws into network architectures or learning system dynamics directly from data using neural operators (also known as emulators). 
The second approach builds on established numerical solvers, which are made differentiable through modern AD frameworks, 
allowing efficient gradient computation with respect to simulation inputs or model parameters in an end-to-end manner.

\subsubsection{Continuous approach: Neural network based methods}

Neural networks have become widely used in physics-based modeling, offering differentiable approximations to  PDE solutions suitable for both forward and inverse problems. 
A prominent approach 
is physics-informed neural networks (PINNs)~\cite{raissi2019physics}, which explicitly enforce governing equations through their loss functions to guide the learning process.
Owing to their architectural flexibility, PINNs have been applied across a wide range of engineering and scientific disciplines~\cite{cuomo2022scientific},
including fluid mechanics~\cite{raissi2020hidden,cai2021physics}, solid mechanics \cite{haghighat2021physics,goswami2020transfer,rao2021physics}, subsurface transport~\cite{he2020physics,du2023modeling}.
However, PINNs often suffer from gradient flow pathologies~\cite{wang2021understanding} and high computational cost due to the evaluation of high-order derivatives~\cite{du2024neural}. These challenges have spurred the development of hybrid neural–numerical approaches, which integrate discrete approximations into the PINN framework to improve accuracy and efficiency~\cite{du2024neural}. 
Despite these advances, the practical application of PINNs in geomechanics and geophysical modeling remains limited, largely due to the added complexity of inelastic material behavior and multiphysics coupling~\cite{yuan2025physics}.

Another notable approach to modeling physical systems with neural networks is to build data-driven surrogates, such as emulators or neural operators, which learn system dynamics directly from observational data. 
These surrogates leverage a range of architectures,
including graph neural networks (GNNs)~\cite{sanchez2020learning}, MeshGraphNet~\cite{pfaff2020learning}, Fourier neural operators (FNO)~\cite{li2020fourier}, and DeepONet~\cite{lu2021learning}.
Recently, a hybrid framework coupling neural operators with the classical FEM has been introduced for solid mechanics~\cite{yin2022interfacing} and  geophysical modeling~\cite{he2023hybrid}, further enhancing the integration of data-driven models with traditional numerical methods.

In Lagrangian (particle-based) settings,
GNN-based simulators are commonly used to model local interactions by dynamically constructing neighborhoods of particles~\cite{allen2022inverse,choi2024inverse}. 
While these methods are very expressive, they often suffer from high computational costs due to repeated neighbor searches and can face scalability challenges. 
To mitigate these issues, a hybrid Eulerian–Lagrangian design has been proposed~\cite{li2023mpmnet,sharabi2024a}
in which particle states are projected onto a background grid via particle-to-grid (P2G) transfers, evolved using grid-based convolutional neural networks (CNNs), and then mapped back to particles using grid-to-particle (G2P) operations. 
This structure avoids explicit neighbor searches and supports more scalable training. 
Despite these innovations, such learned neural simulators remain purely data-driven. They rely heavily on training data, do not explicitly enforce physical laws, and often generalize poorly outside the training distribution,  especially in long-term or multiphysics scenarios.

\subsubsection{Discrete approach: differentiable physics solver}
To circumvent the limitations of neural network-based approaches, increasing attention has been directed toward the second category: 
high-fidelity differentiable physics solvers, also referred to as 
differentiable simulators~\cite{sapienza2024differentiable,newbury2024review}. 
These simulators bridge the gap between classical physics-based modeling and end-to-end learning by constructing solvers that are natively compatible with AD frameworks~\cite{baydin2018automatic}, thereby allowing direct gradient computations through the entire simulation time-stepping process. 
As a result, differentiable simulators provide a physically consistent and scalable foundation for inverse modeling, without requiring pre-training or manual derivation of adjoint equations. 
By supporting gradient-based optimization with respect to material properties, initial conditions, and control inputs, differentiable simulators enable a wide range of applications including system identification~\cite{liang2019differentiable}, trajectory and policy optimization~\cite{hu2019chainqueen, chen2022daxbench}, and material design problems~\cite{xu2021end,xue2023jax,hu2025efficient}.

In this work, we focus specifically on  the differentiable implementation of MPM,
which is inherently well-suited for modeling large deformation processes in geophysical systems.
Traditional MPM implementations are typically developed in imperative programming languages such as C++ \cite{kumar2019scalable}, Fortran \cite{zhang2016material}, or high-level scientific computing platforms like Julia \cite{sinaie2017programming, 
HUO2025107189}. 
While these implementations are efficient and widely used for forward simulations, they are not inherently designed 
to support differentiability or seamless integration with AD-based deep learning frameworks.
Recent efforts have explored differentiable MPM solvers implemented in Taichi~\cite{hu2019taichi}, a domain-specific programming language featuring built-in AD capacities~\cite{hu2019difftaichi}.
Taichi has been successfully employed in  various applications, including physics-based modeling and computer graphics~\cite{hu2019taichi, hu2019difftaichi, shi2024geotaichi},
as well as in tasks involving parameter and system identification~\cite{li2023pac,xu2024differentiableStir}.
Despite these advances in MPM implementations, 
their development specifically tailored for large-scale, differentiable geomechanical modeling remains rare. 
Key limitations of existing approaches include restricted AD flexibility, limited multi-GPU support, and a separation between simulation and machine learning components.

These limitations highlight a critical need for developing a fully differentiable Lagrangian simulator for high-fidelity geomechanical simulations, enabling  a unified framework for scalable forward and inverse modeling workflows. 

\subsection{A learning-based meshfree computational framework}
To address this gap, we propose a general-purpose, Lagrangian meshfree solver built on JAX~\cite{jax2018github}, a modern computational backend specifically tailored for differentiable programming.
JAX integrates automatic differentiation, just-in-time (JIT) compilation, and GPU/TPU acceleration through the accelerated linear algebra (XLA) compiler. Unlike imperative programming frameworks, which typically require explicit control over memory and execution flow, JAX promotes a functional, vectorized programming style, enabling efficient, loop-free simulations fully compatible with gradient-based optimization. 
Additionally, JAX integrates seamlessly with modern deep learning libraries such as Flax~\cite{flax2020github}, Haiku~\cite{haiku2020github}, and Equinox~\cite{kidger2021equinox}, supporting the joint training of physical models and neural networks. 
JAX’s built-in parallelization tools, such as \texttt{vmap} and \texttt{pmap}, further facilitate  efficient multi-GPU computation and batched operations, which are essential for large-scale inverse modeling and data-driven geomechanical workflows.
Recently, JAX has been effectively 
combined with canonical numerical methods, such as FEM~\cite{xue2023jax}, finite difference method~\cite{mistani2023jax}, finite-volume CFD solver~\cite{bezgin2023jax}, SPH~\cite{toshev2024jax} and meshless local Petrov-Galerkin method~\cite{du2024differentiable}, among others.

Leveraging these capabilities, we develop JAX-MPM in this study, a GPU-accelerated, fully differentiable meshfree
framework for large-deformation simulation and inverse modeling in geomechanics and solid mechanics. Implemented entirely in JAX, it unifies simulation, automatic differentiation, and neural networks within a single system, eliminating the need for external toolchains or inter-language bindings required by Taichi-based approaches. 
By combining the physical rigor of MPM with JAX’s flexibility, JAX-MPM enables robust, end-to-end pipelines for forward simulation, gradient-based inversion, and learning-driven optimization in natural hazard analysis and related applications.

We evaluate the effectiveness of JAX-MPM through a diverse set of benchmark problems representative of real-world geomechanical systems. These include both 2D and 3D dam-break scenarios and granular collapses, where the simulator accurately captures large deformations, free-surface evolution, and interactions with rigid boundaries. We also highlight the computational advantages of JAX-MPM through efficient GPU acceleration.
For example, JAX-MPM completes 1000 time steps in approximately 8.2 seconds for a 3D dam-break simulation with 1.9 million particles on a $200 \times 100 \times 80$ grid, and around 22 seconds for a granular cylinder collapse with 2.7 million particles on a $400 \times 400 \times 100$ grid, both using single precision on a single GPU.

For inverse modeling, JAX-MPM introduces a unified PDE-constrained optimization framework that accommodates both Lagrangian (particle-based) and Eulerian (region-based) supervision without modifying the simulation pipeline. This flexible formulation allows the integration of diverse observational inputs, such as sparse particle trajectories or fixed-location velocity sensors. We demonstrate its effectiveness on representative inverse tasks including initial velocity reconstruction and spatially varying friction estimation. These tasks span up to $\mathcal{O}(10^3)$ trainable parameters and $\mathcal{O}(10^4)$ degrees of freedom. In all cases, JAX-MPM achieves accurate, physically consistent reconstructions using only partial observational data, underscoring its potential for real-world geohazard applications.
To the best of our knowledge, 
this is the first attempt to employ a differentiable Lagrangian physics solver for 
large-scale inverse modeling in geophysical flow applications.

\subsection{Aims and structure}

The main contributions of this work are summarized as follows: 1) develop JAX-MPM, a \textit{fully differentiable GPU-accelerated meshfree solver} for high-fidelity geomechanical simulations; 2) introduce a \textit{unified inverse modeling framework} supporting both Lagrangian (particle-based) and Eulerian (region-based) data, enabling flexible and scalable inference of physical parameters from diverse observational sources; and 3) demonstrate the potential of JAX-MPM for hybrid physics–machine learning workflows, facilitating end-to-end differentiable pipelines for parameter estimation and field reconstruction.

The remainder of the paper is organized as follows. Section~\ref{sec:Physical} introduces the physical modeling and numerical formulation of the material point method employed in JAX-MPM. Section~\ref{sec:inverse_sec3} details the implementation of scalable inverse modeling within the JAX-MPM framework. Section~\ref{sec:result} presents forward simulation results for representative benchmark problems. Section~\ref{sec:Inverse_dam} demonstrates inverse modeling capabilities, including parameter estimation and field reconstruction. Finally, Section~\ref{sec:conclusion} concludes the paper with a summary and future directions.

%% file: Preliminaries.tex
\section{Preliminary}\label{sec:Physical}
In this section, we  briefly review the governing equations and  explicit time integration scheme underlying the proposed differentiable meshfree framework based on MPM, as detailed  in Sections \ref{sec:formulation} and   \ref{sec:Implement}. 
We then introduce several classical constitutive material models employed in this study (Sections \ref{sec:constitutive}).

\subsection{Governing equations}\label{sec:formulation}
The balance of linear momentum in the Lagrangian (material) frame, defined over a domain \( \Omega \subset \mathbb{R}^d \), where \( d \) is the spatial dimension and \( t \in [0, T] \), is given by:
\begin{equation}
\left\{
\begin{aligned}
& \rho \dot{\bm{v}} = \nabla \cdot \bm{\sigma} + \rho \bm{b}, \\
& \bm{\sigma} \cdot \bm{n} = \bar{\bm{t}} \quad \text{on } \Gamma_N, \\
& \bm{u} = \bar{\bm{u}} \quad \text{on } \Gamma_D,
\end{aligned}
\right.
\label{eq:governing}
\end{equation}
where \( \rho \) denotes the material density, \( \bm{b} \) is the body force per unit mass, and \( \bm{\sigma} \) is the Cauchy stress tensor. The boundary \( \partial \Omega \) is partitioned into Dirichlet (\( \Gamma_D \)) and Neumann (\( \Gamma_N \)) portions, corresponding to prescribed displacements $\bar{\bm{u}}$ and tractions $\bar{\bm{t}}$, respectively. The velocity field is defined as \( \bm{v} = \dot{\bm{u}} \), where \( \bm{u} \) is the displacement field, and the superposed dot denotes the material derivative. 

The weak form of Eq.~\eqref{eq:governing} is derived by multiplying with a vector-valued test function \( \bm{w} \), integrating over the current configuration, and applying the divergence theorem:
 
\begin{equation}
\int_{\Omega} \bm w \cdot \rho \dot{\bm v} \mathrm{~d}\Omega=-\int_{\Omega} \nabla \bm w: \bm \sigma \mathrm{~d} \Omega+\int_{\Omega} \bm w \cdot \rho \boldsymbol{b} \mathrm{~d} \Omega +\int_{\Gamma_{{N}}} \bm w \cdot \bar{\bm t} \mathrm{~d} \Gamma
\label{eq:weak}
\end{equation}
where \( \bm{w} \) vanishes on \( \Gamma_D \). 

\begin{figure}
    \centering
    \includegraphics[width=0.65\linewidth]{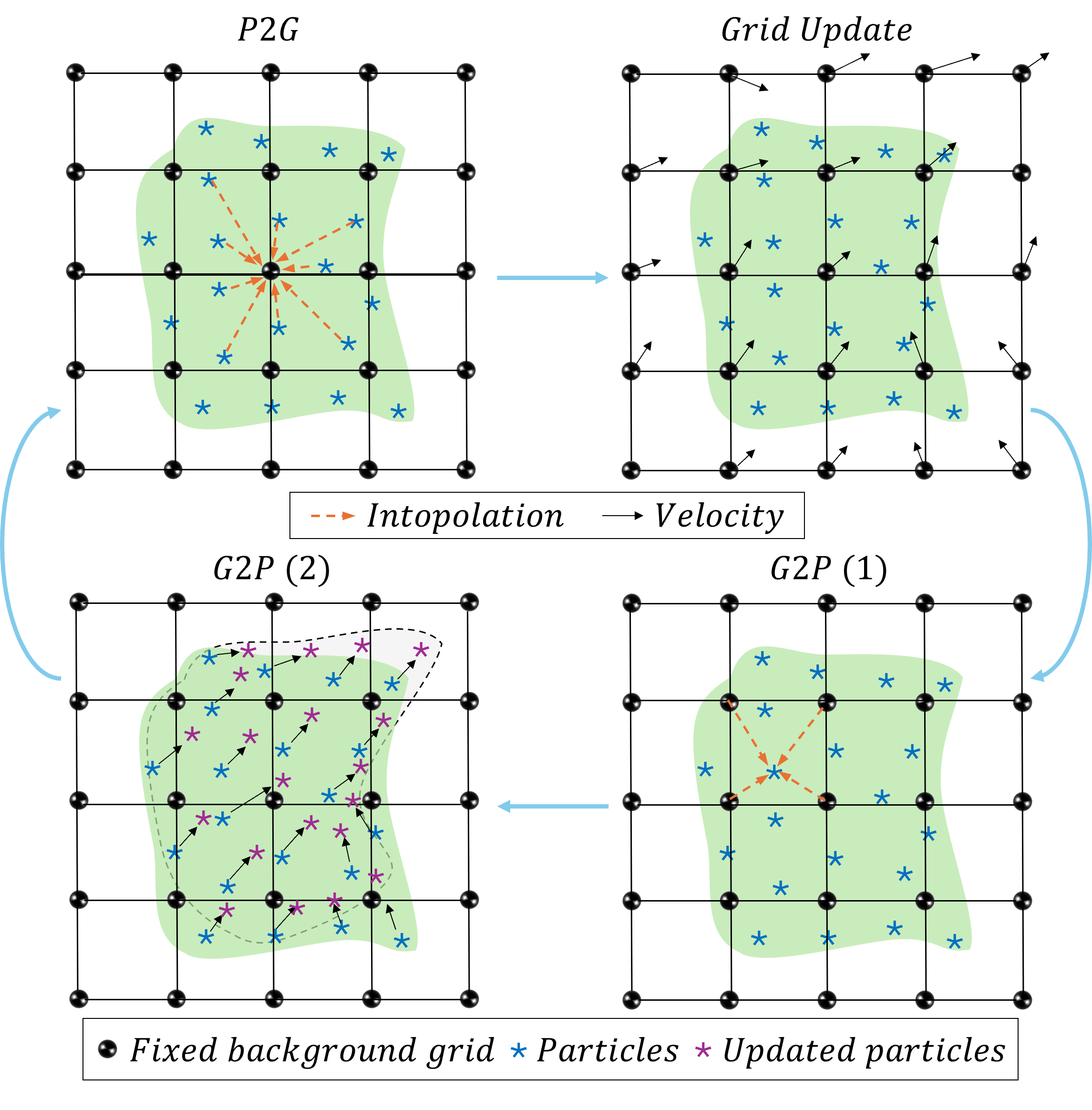}
    \caption{Illustration of the material point method (MPM) procedure. The procedure consists of three stages: P2G transfer, grid-based updates, G2P transfer. The G2P operation includes grid-to-particle interpolation, particle-based updates, and grid reset. Black dots denote the fixed background grid, blue stars represent particles, and purple stars indicate updated particles. Dashed orange arrows illustrate interpolation contributions, while solid black arrows indicate velocity vectors.}
    \label{fig:mpm_plot}
\end{figure}

Following the standard MPM framework~\cite{sulsky1994particle, sulsky1995application} (illustrated in Fig. \ref{fig:mpm_plot}), the continuum domain is discretized into a collection of Lagrangian material points (particles) located at positions \( \bm{x}_p \), \( p = 1, \dots, N_p \). These particles  move freely over a fixed Eulerian background grid, whose nodes are located at $ \bm{x}_i $, \( i = 1, \dots, N_g \). 
Throughout this paper, the subscript $p$ refers to material points, while the subscript 
$i$ refers to grid nodes.

Each material point is associated with a subdomain \( \Omega_p \subset \Omega \), and it carries the physical properties of the continuum, such as mass, volume, velocity, and stress.
This particle-based representation allows the weak form of the momentum balance equation~\eqref{eq:weak} to be expressed  as a sum over all material points:
\begin{equation} 
\sum_{p} \int_{\Omega_p} \bm w \cdot \rho \dot{\bm v} \mathrm{~d} \Omega=- \sum_{p} \int_{\Omega_p} \nabla \bm w: \bm \sigma \mathrm{~d} \Omega+ \sum_{p} \int_{\Omega_p} \bm w \cdot \rho \boldsymbol{b} \mathrm{~d} \Omega+
\sum_{p} \int_{\Gamma_{{N},p}} \bm w \cdot 
\overline{\boldsymbol{t}} \mathrm{~d} \Gamma
\label{eq:discrete}
\end{equation} 
where the index $p$ denotes the 
material points in the domain, and $\Gamma_{{N},p}$ denotes the portion of the Neumann boundary $\Gamma_{N}$ intersecting the subdomain $\Omega_p$.
It is worth noting that the conservation of mass is inherently satisfied in the MPM formulation and is therefore omitted from the present derivation \cite{sulsky1994particle}.


The velocity $\bm v (\bm x)$ and test function $\bm w (\bm x)$ are discretized on the background grid as linear combinations of the nodal shape functions \( \phi_i(\bm{x}) \):
\begin{equation}
\bm v=\sum_{i} \phi_i(\bm x) \bm v_i, \quad \bm w=\sum_{i} \phi_i(\bm x) \bm w_i
\label{eq:fe}
\end{equation}  
where the index $i=1,...,N_g$ denotes the grid nodes,
\( \phi_i(\bm{x}) \) denotes the shape function associated with node \( i \) with a compact support,
and $\bm v_i$ and $\bm w_i$ are the corresponding nodal velocity and nodal test function, respectively. 
In this study, the multidimensional shape function \( \phi_i(\bm{x}) \) is constructed using a tensor product of one-dimensional B-splines:
\begin{equation}
\phi_i(\bm{x}) = \prod_{\alpha=1}^d \bar{\phi}\left( \frac{x_\alpha - x_{i\alpha}}{\Delta h} \right)
\end{equation}
where \( \Delta h \) is the uniform grid spacing, and 
the quadratic B-spline basis function $\bar{\phi}$ is defined as~\cite{steffen2008analysis}:
\begin{equation}
\begin{aligned}
\bar{\phi}(\xi) & = \begin{cases}-|\xi|^2+\frac{3}{4} & 0 \leq|\xi|<\frac{1}{2} \\
\frac{1}{2}|\xi|^2-\frac{3}{2}|\xi|+\frac{9}{8} & \frac{1}{2} \leq|\xi|<\frac{3}{2} \\
0 & \frac{3}{2} \leq|\xi|\end{cases} \\
\end{aligned}
\end{equation}

Substituting Eq.~\eqref{eq:fe} into the discrete weak form \eqref{eq:discrete} and considering the arbitrariness of nodal test functions \( \bm{w}_i \), we obtain the semi-discrete form:

\begin{equation}
\sum_{j=1}^{N_g} m_{ij} \dot {\bm v}_j=\bm f_i^{\mathrm{int}}+\bm f_i^{\mathrm{ext}}
\label{eq:discre_eq}
\end{equation}
where
\begin{equation}
\begin{aligned}
m_{i j} & = \sum_{p} \int_{\Omega_p} \phi_i \phi_j \rho \mathrm{~d} \Omega= \sum_{p} m_p \phi_{i p} \phi_{j p} \\
\bm f_i^{\mathrm{int}} & =- \sum_{p} \int_{\Omega_p} \nabla \phi_i \cdot \bm \sigma \mathrm{~d} \Omega=- \sum_{p} V_p \nabla \phi_{i p} \cdot \bm \sigma_p \\
\bm f_i^{\mathrm{ext}} & = \sum_{p}\left[\int_{\Omega_p} \phi_i \rho \bm b \mathrm{~d} \Omega \right]+
\sum_{p} \int_{\Gamma_{{N},p}}
\phi_i \bar{\bm t} \mathrm{~d} \Gamma= \sum_{p} \phi_{i p} m_p \bm b_p+\bar{\bm f}_i^t 
\end{aligned}
\label{eq:semi}
\end{equation}
where \( \phi_{i p} = \phi_i(\bm x_p)  \), $V_p$, \( m_p \) and $\bm{\sigma}_p$ are the volume, mass and stress tensor associated with particle \( p \), respectively, \( m_{ij} \) is the consistent mass matrix, and $\bar{\bm f}_i^t$ is the nodal force contributed by the traction $\bar{\bm{t}}$. 

To avoid the costly inversion of the consistent mass matrix~\cite{sulsky1994particle}, we adopt a lumped mass matrix, which retains only the nonzero diagonal entries, $m_i = \sum_{p} m_p \phi_{ip}$.
With this lumped mass matrix, application of forward Euler time integration to the semi-discrete momentum equation \eqref{eq:discre_eq} yields the following explicit time-stepping scheme:
\begin{equation}
m_i^n \frac{\bm v_i^{n+1}-\bm v_i^n}{\Delta t} = \bm{f}_i^{\mathrm{int}, n} + \bm{f}_i^{\mathrm{ext}, n} : =\bm f_i^n
\label{eq:semi_eq}
\end{equation}
where the superscript \( n \) denotes the current time step, \( \Delta t = t^{n+1} - t^n \) is the time increment,
and \( \bm{f}_i^n = \bm{f}_i^{\mathrm{int}, n} + \bm{f}_i^{\mathrm{ext}, n} \) is the total force acting on node \( i \), consisting of internal and external contributions. 
To ensure the numerical stability of the explicit scheme, $\Delta t$ is 
chosen according to the Courant–Friedrichs–Lewy (CFL) condition~\cite{courant1967partial}.

\subsection{Implementation of material point method}\label{sec:Implement}

As shown in Fig. \ref{fig:mpm_plot}, 
each time step in the MPM solution procedure consists of three computational stages:
particle-to-grid (P2G) transfer, grid-based updates, and grid-to-particle (G2P) transfer. The computation procedures at three stages are summarized below.

\subsubsection*{Particle-to-grid (P2G) transfer}

In the P2G phase, particle information is projected onto the background grid using interpolation kernels. The mass at grid node \( i \), consistent with the lumped mass matrix, is computed via a weighted sum over particles
\begin{equation}
    m_i^n = \sum_p m_p \phi_{ip}^n.
    \label{eq:update_m}
\end{equation}
The momentum contributions from particles are aggregated to compute the grid velocity:
\begin{equation}
    \bm{v}_i^n = \frac{1}{m_i^n} \sum_p m_p \phi_{ip}^n \bm{v}_p^n.
\end{equation}
The nodal force vector $\bm{f}_i^n$ consists of the  internal and external force vectors defined in Eq. \eqref{eq:semi}:
\begin{equation}
    \bm{f}_i^n = -\sum_p V_p^n \nabla \phi_{ip}^n \cdot \bm{\sigma}_p^n 
    + \sum_p \phi_{ip}^n m_p \bm{b}_p 
    + 
    \bar{\bm f}_i^t,
\end{equation}
where \( V_p^n \) and \( \bm{\sigma}_p^n \) are the volume and stress of particle \( p \), \( \bm{b}_p \) is the body force per unit mass, and \( \bar{\bm f}_i^t \) represents the nodal
force contributed by the traction, as defined in Eq.~\eqref{eq:semi}.
The constitutive model to compute  $\bm \sigma_p^n$ is given in Section \ref{sec:constitutive}.

\subsubsection*{Grid-based updates}

With the nodal representation of mass and force at node $i$, the grid acceleration is 
given as:
\begin{equation}
    \bm{a}_i^n = \frac{\bm{f}_i^n}{m_i^n}.
\end{equation}
The grid velocity is then updated using an explicit time integration scheme:
\begin{equation}
    \bm{v}_i^{n+1} = \bm{v}_i^n + \Delta t \bm{a}_i^n.
\end{equation}
In explicit MPM, Dirichlet conditions are typically imposed by overwriting boundary‐node velocities. However, with higher-order kernels (e.g., quadratic B-splines), particle supports are truncated at the domain edge, which breaks partition of unity and leads to spurious stress oscillations and potential instabilities \cite{de2020material}. To address this issue, two approaches can be used: (i) adding \textit{ghost cells} so that boundary-adjacent particles retain a complete stencil \cite{de2020material}, thereby restoring partition of unity (this approach is adopted in the present work); and (ii) applying a \textit{local kernel-correction} near boundaries to enforce the partition of unity \cite{nakamura2023taylor}.

\subsubsection*{Grid-to-particle (G2P) transfer}\label{sec:g2p_trans}

Following the update of grid velocities, information is transferred back to the particles to advance the simulation. JAX-MPM supports several established transfer schemes, including fluid-implicit particle (FLIP)~\cite{brackbill1986flip}, particle-in-cell (PIC)~\cite{harlow1964particle}, affine-PIC (APIC) \cite{jiang2015affine} and Taylor-PIC (TPIC) \cite{nakamura2023taylor}. In this section, we describe the commonly used FLIP and PIC methods; additional variants are detailed in~\ref{appendix:pic}.

The FLIP scheme updates the particle velocity using the grid acceleration:
\begin{equation}
    \bm{v}_p^{n+1} = \bm{v}_p^n + \Delta t \sum_i \phi_{ip}^n \bm{a}_i^n.
\end{equation}
Alternatively, the PIC scheme directly interpolates the updated grid velocity
\begin{equation}
    \bm{v}_p^{n+1} = \sum_i \phi_{ip}^n \bm{v}_i^{n+1}.
\end{equation}
PIC replaces the particle velocity with the interpolated grid velocity, introducing numerical dissipation into the system~\cite{harlow1964particle}. In contrast, FLIP only adds the increment of grid velocity, which preserves particle-level quantities and reduces numerical dissipation, but may introduce noise and instability~\cite{brackbill1986flip, de2020material}. To balance these effects, a blended scheme \cite{zhu2005animating} can be used.

The particle position is updated using the nodal velocity field:
\begin{equation}
    \bm{x}_p^{n+1} = \bm{x}_p^n + \Delta t \sum_i \phi_{ip}^n \bm{v}_i^{n+1},
\end{equation}
and the velocity gradient at each particle (e.g., used to update the stress tensor) is evaluated by:
\begin{equation}
    \nabla \bm{v}_p^{n+1} = \sum_i \bm{v}_i^{n+1} \otimes \nabla \phi_{ip}^n.
    \label{eq:update_v_gradient}
\end{equation}

At the end of the G2P stage, the quantities, such as velocity, are reset before P2G stage for next time step.


\vspace{0.2em}


\subsubsection*{Remark on stress update schemes}
A critical aspect of MPM implementation is the choice of stress update scheme, which defines when stresses are evaluated relative to the solution of the momentum equation. 
Two common stress update schemes in MPM are \emph{update stress last (USL)} and \emph{update stress first (USF)}~\cite{bardenhagen2002energy}. USL updates stresses after computing internal forces and advancing velocities, offering better stability and modularity. USF updates stresses earlier, during the P2G step, which can improve energy conservation but may be less stable in dynamic problems. 
The implementation described above follows the USL approach.


\subsection{Constitutive model}\label{sec:constitutive}

JAX-MPM supports constitutive models tailored for large-deformation geophysical flows. 
In this study, we consider two representative materials: 
(1) Newtonian fluids used in the dam-break scenarios, and  
(2) elastoplastic solids governed by a Drucker–Prager (D–P) yield criterion with a tensile cutoff, commonly employed for granular media.
The stress update procedures for each material model are detailed below.

\begin{itemize}

\item  For the Newtonian fluid model, the Cauchy stress tensor \( \bm{\sigma}_p^{n+1} \) is updated as

\begin{equation}
    \boldsymbol{\sigma}_p^{n+1} = -p_p^{n+1} \boldsymbol{I} - \frac{2}{3} \mu \operatorname{tr}(\Delta \boldsymbol{d}_p) \boldsymbol{I} + 2 \mu \Delta \boldsymbol{d}_p,
    \label{eq:stress_strain}
\end{equation}  
where $\bm d_p$ represents the rate of deformation tensor, and $\Delta \bm d_p=\frac{1}{2}(\nabla \bm v_p^{n+1}+\nabla^{\top} \bm v_p^{n+1}) \Delta t$, \( \bm{I} \) is the identity tensor, and \( \mu \) is the dynamic viscosity. 

To model incompressibility, we adopt a weakly compressible formulation, where the pressure is computed using an artificial equation of state (EOS) \cite{koh2012new,li2014sloshing, molinos2023derivation}

\begin{equation}
    p_p^{n+1} =  c_n^2 (\rho_p^{n+1}-\rho_0)
    \label{eq:eos}
\end{equation}
where \( c_n \) is the numerical sound speed, \( \rho_0 \) is the reference density of fluid, and \( \rho_p^{n+1} \) is the updated fluid density at time step \( (n+1) \), computed as $ \rho_p^{n+1}=\rho_p^{n} /(1+\operatorname{det}(\Delta \bm d_p))$.

\item For granular flow, we adopt a non-associated Drucker–Prager (D-P) plasticity model with a tension cutoff \cite{huang2015large, nguyen2020effects}. 
To maintain objectivity under large deformation, we employ the Jaumann objective stress rate. The stress update over a time step is written as:
\begin{equation}
\bm \sigma_p^{n+1} =\bm \sigma_p^n+ \bm \sigma_p^{\nabla J} \Delta t + \left( \Delta \omega_p \cdot \bm \sigma_p^n + \bm \sigma_p^n \cdot \Delta \omega_p^{T} \right)
\label{eq:dp_stress}
\end{equation}
where $\bm \sigma_p^{\nabla J}$ denotes the Jaumann rate of Cauchy stress, and \( \Delta \bm{\omega}_p \) is the vorticity increment, defined as  $\Delta \bm{\omega}_p = \frac{1}{2} \left[ \nabla \bm{v}_p^{n+1} - \left( \nabla \bm{v}_p^{n+1} \right)^{\top} \right] \Delta t$. Note that a return mapping algorithm is applied to enforce the Drucker–Prager yield criterion during plastic correction \cite{huang2015large, drucker1952soil} on the Jaumann stress component. The implementation details, including both shear and tensile failure modes, are provided in ~\ref{appendix:dp}.
\end{itemize}

%% file: approximation.tex

\section{JAX-MPM: Differentiable meshfree solver for forward and inverse modeling}\label{sec:inverse_sec3}

JAX-MPM is implemented within the JAX framework~\cite{jax2018github}, a modern system for differentiable programming with XLA-based GPU/TPU acceleration. This design enables high-performance forward particle-based simulations while providing efficient gradient-based optimization capabilities for inverse modeling and data assimilation. Moreover, the solver can integrate seamlessly  with network–based machine learning (ML) models, facilitating hybrid physics–ML workflows.

\subsection{Gradient computation by automatic differentiation}
\label{sec:grad_ad}

JAX-MPM formulates the physical simulation as a composition of differentiable operators, allowing 
reverse-mode automatic differentiation to propagate gradients through entire trajectories.

Let a set \( \mathbf{S}^t = \left\{ \bm{x}_p^t, \bm{v}_p^t, \bm{F}_p^t, \rho_p^t \right\}_{p=1}^{N_p} \in \mathcal{S}\)
denote the collection of structured simulation states at time step \( t \), 
where each material point is characterized by its position \( \bm{x}_p^t \in \mathbb{R}^d \), velocity \( \bm{v}_p^t \in \mathbb{R}^d \), deformation gradient \( \bm{F}_p^t \in \mathbb{R}^{d \times d} \), and density \( \rho_p^t \in \mathbb{R} \). The space $\mathcal{S}$ denotes the appropriate vector space for these simulation states. 
We note that additional 
quantities, such as stress \( \bm{\sigma}_p^t \in \mathbb{R}^{d \times d} \) and equivalent plastic strain \( \varepsilon_{eq,p}^t \in \mathbb{R} \), may be included as needed but are omitted here for simplicity.

The state update from time $t$ to $t+1$ in the JAX-MPM solver (see Fig. \ref{fig:mpm_plot} and Eqs. \eqref{eq:update_m}–\eqref{eq:update_v_gradient}) 
is defined by the mapping:
\begin{equation}\label{eq:mapping}
\mathbf{S}^{t+1} = \Phi(\mathbf{S}^t; \boldsymbol{\alpha}),\qquad
\Phi:\mathcal{S}\times\mathbb{R}^m\to\mathcal{S},
\end{equation}
where $\boldsymbol{\alpha}\in\mathbb{R}^m$ denotes the set of model parameters governing the dynamics (e.g., material properties, friction, boundary/initial conditions). As in Section~\ref{sec:Implement}, the update operator $\Phi$ consists of three stages: {particle-to-grid} transfer (P2G), {grid-based updates} (GU), and {grid-to-particle} transfer (G2P), such that
\(
\Phi = \Phi_{\text{G2P}} \circ \Phi_{\text{GU}} \circ \Phi_{\text{P2G}}.
\)
The  forward simulation trajectory over $N_t$ steps is thus expressed as: 
\begin{equation}
\mathbf{S}^0 
\xrightarrow{\Phi(\boldsymbol{\alpha})} 
\mathbf{S}^1 
\xrightarrow{\Phi(\boldsymbol{\alpha})} 
 \cdots  
\xrightarrow{\Phi(\boldsymbol{\alpha})} 
\mathbf{S}^{N_t}
\label{eq:trajectory}
\end{equation}

Over a complete trajectory with $N_t$ time steps, the ensemble of simulation states is given by \( \mathcal{X} = \{ \mathbf{S}^t \}_{t=0}^{N_t} \). Accordingly, the discrete dynamics equations \eqref{eq:trajectory} on $\mathcal{X}$ can be  expressed compactly as:
\begin{equation}
\mathcal{J}(\mathcal{X}, \boldsymbol{\alpha}) = \mathbf{0}.
\end{equation}
where $\mathcal{J}$ denotes the operator that enforces the physical constraints.

By making the time-stepping operator \( \Phi \) \textit{differentiable} within the JAX-MPM framework,
efficient gradient computation with respect to the physical parameters \( \boldsymbol{\alpha} \) can be obtained via reverse-mode automatic differentiation.
Consider a state variable of interest extracted from the full state set $\mathbf{S}$ at time step \( (t+1) \), 
$ \boldsymbol{z}^{t+1} \in \mathbf{S}^{t+1}$,
for example, velocity \( \boldsymbol{z}^{t+1} := \boldsymbol{v}^{t+1}\). 
Its sensitivity with respect to the parameters \( \boldsymbol{\alpha} \) can be evaluated recursively through time using the chain rule:
\begin{equation}
\frac{d \boldsymbol z^{t+1}}{d\boldsymbol{\alpha}} = 
\frac{\partial \Phi_z^t}{\partial \mathbf{S}^t} \cdot \frac{d \mathbf{S}^t}{d\boldsymbol{\alpha}} + 
\frac{\partial \Phi_z^t}{\partial \boldsymbol{\alpha}},
\label{eq:dzda}
\end{equation}
where the subscript $z$ in $\Phi_{\boldsymbol{z}}^t$ indicates that the component $\boldsymbol{z}$ is extracted from the output of the time-stepping operator evaluated using input $\mathbf{S}^t$.
The recursion in \eqref{eq:dzda} continues backward in time via
Eq. \eqref{eq:mapping}, e.g., 
\begin{equation}
  \frac{d \mathbf{S}^t}{d\boldsymbol{\alpha}} = 
\frac{\partial \Phi^{t-1}}{\partial \mathbf{S}^{t-1}} \cdot \frac{d \mathbf{S}^{t-1}}{d\boldsymbol{\alpha}} + 
\frac{\partial \Phi^{t-1}}{\partial \boldsymbol{\alpha}}.
\end{equation}

In practice, this recursion is handled automatically by JAX’s autodiff engine. 
For example, consider a scalar quantity $z^{t+1}$ that depends on $\boldsymbol{\alpha}$.
Its gradient can be obtained by invoking \texttt{jax.grad} 
\begin{equation}
\frac{d z^{t+1}}{d\boldsymbol{\alpha}}
= \texttt{jax.grad}(z^{t+1})(\boldsymbol{\alpha}).
\end{equation}
JAX constructs the computation graph during the forward pass and subsequently evaluates gradients by traversing it backward using reverse-mode vector–Jacobian products, thereby eliminating the need for explicit recursive implementation.


\subsection{PDE-constrained optimization for inverse modeling}
\label{sec:inverse_optim}

Inverse modeling plays a central role in geophysical hazard analysis, where key physical parameters, such as initial velocities or spatially varying friction, cannot be directly observed but significantly influence flow behavior and deposition in events like landslides, dam breaks, and debris flows. 
We formulate parameter estimation as a {PDE-constrained optimization problem}, in which the unknown parameters \( \boldsymbol{\alpha} \) are inferred by minimizing the discrepancy between simulated and observed data, subject to physical governing  equations implemented within JAX-MPM.

To this end, we introduce a differentiable output mapping operator \( \mathcal{Q} \) that extracts simulated quantities corresponding to observed state data 
\( \boldsymbol{z}_{l,t}^{\text{obs}} \in \mathbb{R}^d \) 
at spatial index \( l \) and time \( t \).
This operator provides a unified interface that accommodates both Lagrangian (particle-based) and Eulerian (region-based) forms of supervision.
The simulated counterpart at index \( l \) is denoted \( \mathcal{Q}_l(\boldsymbol{z}^{t}) \), where \( \boldsymbol{z}^{t} \in \mathbf{S}^{t} \) represents the model state associated with the observation.

\begin{itemize}

\item In the Lagrangian setting, \( l \) denotes a tracked particle, 
and the operator simply returns its state:
\begin{equation}
\mathcal{Q}_l(\boldsymbol{z}^{t}) = 
[\boldsymbol{z}^{t}]_{l}.
\end{equation}

\item In the Eulerian setting, \( l \) corresponds to a fixed spatial coordinate \( \mathbf{x}_l \), associated with a compact surrounding region \( \Omega_l \subset \mathbb{R}^d \) centered at \( \mathbf{x}_l \). The simulated output is defined as the average over particle states within this region:
\begin{equation}
\mathcal{Q}_l(\boldsymbol{z}^{t}) = \frac{1}{|\mathcal{P}_l^t|} \sum_{p \in \mathcal{P}_l^t} \bm z_p^{t},
\end{equation}
where $\mathcal{P}_l^t = \{ p \mid \bm{x}_p^t \in \Omega_l\}$ denotes the set of material points located within the Eulerian region \( \Omega_l \).
\end{itemize}

\begin{figure}[htbp]
    \centering
    \includegraphics[width=1\linewidth]{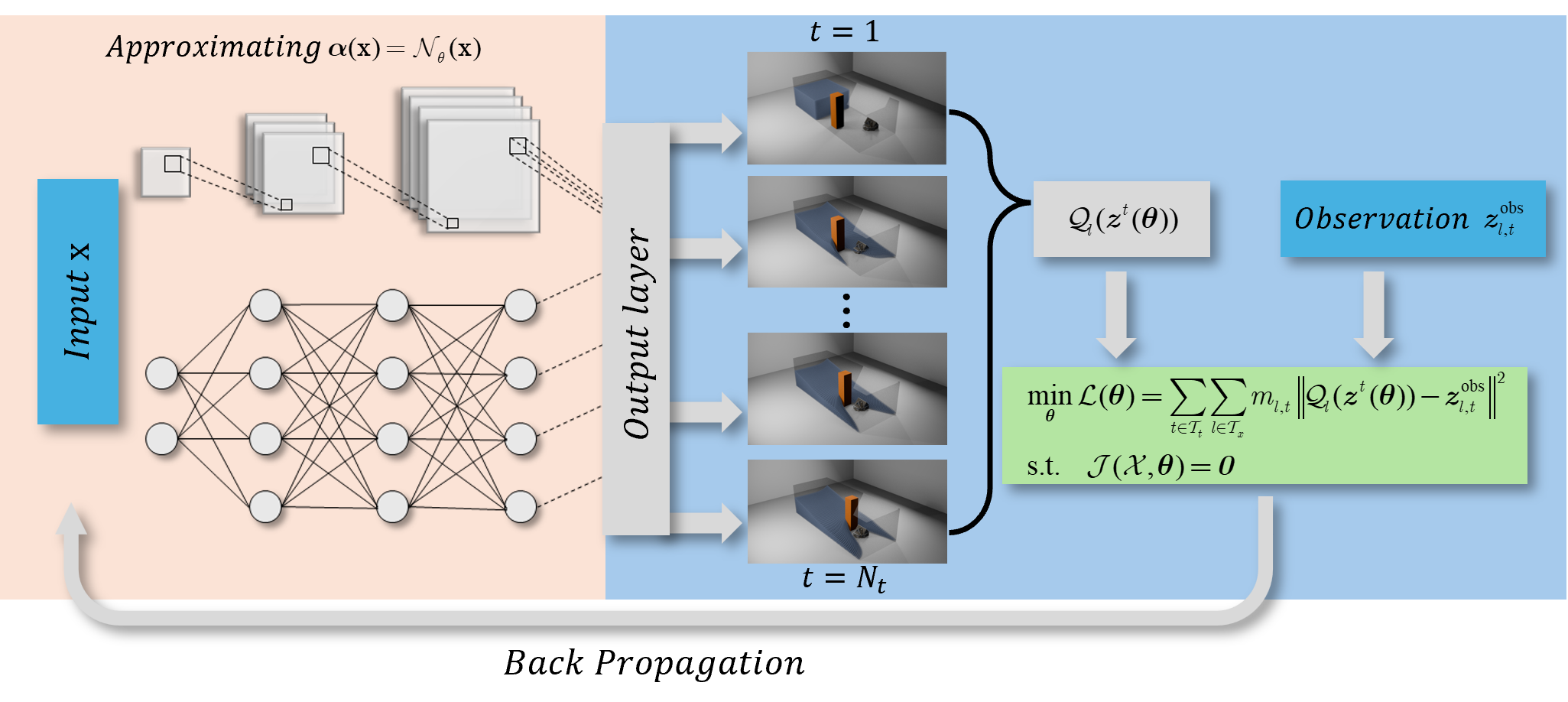}
    \caption{Schematic illustration of the inverse modeling framework in JAX-MPM. A neural network \( \mathcal{N}_\theta(\mathbf{x}) \) approximates the unknown spatially varying parameter field \( \alpha(\mathbf{x}) \), which serves as input to the differentiable simulator. The simulation generates state trajectories \( \boldsymbol{z}^t(\theta) \), from which observable quantities \( \mathcal{Q}_l(\boldsymbol{z}^t(\theta)) \) are extracted and compared against observed data \( \boldsymbol{z}^{\mathrm{obs}}_{l,t} \). The loss is computed over selected space-time points and backpropagated through the entire pipeline via automatic differentiation to update the neural network parameters \( \theta \).}

    \label{fig:framework}
\end{figure}

The generic inverse problem is depicted in Fig. \ref{fig:framework} and can be formulated as:
\begin{align}\label{eq:inverse_loss}
\begin{split}
& \min_{\boldsymbol\alpha} \ \mathcal{L}(\boldsymbol{\alpha}) = \sum_{t \in \mathcal{T}_t} \sum_{l \in \mathcal{T}_x} m_{l,t} \left\| \mathcal{Q}_l(\boldsymbol{z}^t(\boldsymbol{\alpha})) - \bm{z}_{l,t}^{\text{obs}} \right\|^2, \\
& \text{s.t.} \quad \mathcal{J}(\mathcal{X}, \boldsymbol{\alpha}) = \mathbf{0}.
\end{split}
\end{align}
where the discrete physics $\mathcal{J}(\mathcal{X},\boldsymbol{\alpha})=\boldsymbol{0}$ are enforced implicitly by the differentiable time-stepping operator $\Phi$ in Eq.~\eqref{eq:mapping} along the trajectory $\mathcal{X}=\{\mathbf{S}^t\}_{t=0}^{N_t}$. 
The binary mask \( m_{l,t} \in \{0, 1\} \) indicates the availability of ground-truth data at spatial index \( l \) and time \( t \).
$\boldsymbol{z}_{l,t}^{\text{obs}}\in\mathbb{R}^d$ denotes the observed quantity at spatial index $l$ and time $t$.
The set of observed time indices is $\mathcal{T}_t\subseteq\{0,\dots,N_t\}$, where $N_t$ is the total number of time steps.
The set of spatial indices is $\mathcal{T}_x$, which depends on the supervision type:
\begin{equation}
\left\{\begin{array}{l}
\mathcal{T}_x \subseteq\left\{1, \ldots, N_p\right\} \quad \text { (Lagrangian; particles) }, \\
\mathcal{T}_x \subseteq\left\{1, \ldots, N_m\right\} \quad \text { (Eulerian; monitor regions) },
\end{array}\right.
\end{equation}
with $N_p$ denoting the number of particles and $N_m$  the number of monitor regions. 

In classical adjoint-based PDE optimization, gradients w.r.t.\ \(\boldsymbol{\alpha}\) require deriving and solving backward recurrences. In our setting, the loss gradient decomposes as
\begin{equation}
\frac{d \mathcal{L}}{d \boldsymbol{\alpha}} =
\sum_{t \in \mathcal{T}_t} \sum_{l \in \mathcal{T}_x}
m_{l,t}\,\bigl( \mathcal{Q}_l(\boldsymbol{z}^t) - \boldsymbol{z}_{l,t}^{\text{obs}} \bigr)^{\top}
\frac{d \mathcal{Q}_l(\boldsymbol{z}^t)}{d \boldsymbol{\alpha}},
\quad
\frac{d \mathcal{Q}_l(\boldsymbol{z}^t)}{d \boldsymbol{\alpha}} =
\frac{\partial \mathcal{Q}_l}{\partial \boldsymbol{z}^t}\,
\frac{d \boldsymbol{z}^t}{d \boldsymbol{\alpha}},
\end{equation}
with \(d\boldsymbol{z}^t/d\boldsymbol{\alpha}\) obtained recursively via Eq.~\eqref{eq:dzda}.
Rather than implementing adjoint solvers manually, JAX-MPM leverages reverse-mode automatic differentiation (AD), as detailed in Section~\ref{sec:grad_ad}:
\begin{equation}
\frac{d \mathcal{L}}{d \boldsymbol{\alpha}} = \texttt{jax.grad}(\mathcal{L})(\boldsymbol{\alpha}).
\end{equation}

We remark that the parameter field \( \boldsymbol{\alpha} \) can take several forms, depending on the inverse problem. It may represent (i) discrete values specified at particular time steps (e.g., an initial velocity), (ii) trajectory-wide constants (e.g., basal friction or material parameters), or (iii) a field varying in space and/or time, i.e., \( \boldsymbol{\alpha}=\boldsymbol{\alpha}(\mathbf{x},t) \).

When \( \boldsymbol{\alpha} \) varies spatially or temporally, direct optimization over high-dimensional fields is costly. We therefore parametrize the field with a light neural network
\( \mathcal{N}_\theta \) with weights \( \boldsymbol{\theta}\in\mathbb{R}^{N_\theta} \) (see Fig.~\ref{fig:framework}), such that
\begin{equation}
    \boldsymbol{\alpha}(\mathbf{x},t) \approx  \boldsymbol{\alpha}(\mathbf{x},t; \boldsymbol{\theta}) \;:=\; \mathcal{N}_\theta(\mathbf{x},t)
\label{eq:nn_apprx}
\end{equation}
This approach provides a smooth and generalized representation of the parameter field for efficient learning~\cite{he2020physics,du2024differentiable}. 
In this setting, the inverse problem in Eq. \eqref{eq:inverse_loss} optimizes \( \boldsymbol{\theta} \) rather than \( \boldsymbol{\alpha} \); equivalently, one solves \( \min_{\boldsymbol{\theta}}\mathcal{L}(\boldsymbol{\theta}) \) with the constraint \( \mathcal{J}(\mathcal{X},\boldsymbol{\alpha}(\boldsymbol{\theta}))=\boldsymbol{0} \). 
Gradients with respect to either \( \boldsymbol{\alpha} \) or \( \boldsymbol{\theta} \) are computed via reverse-mode automatic differentiation through the simulator and the network.

Numerical studies demonstrating inverse modeling using both Lagrangian and Eulerian observations, as well as neural network parameterizations, are presented in Section~\ref{sec:Inverse_dam}.


\subsection{Software implementation details}\label{sec:Software}

The gradient computations described in the previous section rely on differentiating through unrolled simulation trajectories, which can be both computationally intensive and memory-demanding. 
To support large-scale differentiable simulation, JAX-MPM is implemented on top of the JAX framework, which provides just-in-time (JIT) compilation, automatic differentiation (AD), and GPU acceleration. These capabilities are crucial for achieving high computational efficiency and enabling gradient-based optimization. In the context of inverse modeling of dynamic systems, simulation steps must be executed repeatedly over long time horizons, making both computational speed and memory efficiency critical. To address these demands, JAX-MPM integrates two core strategies: (1) computational acceleration through kernel fusion and loop-level vectorization, and (2) memory optimization via checkpointing and segmented backpropagation. These components are described in the following sections.



\subsubsection{Computational speedup using JIT and loop optimization} \label{sec:speedup}

As summarized in Section~\ref{sec:Implement}, MPM simulations consist of three core computational stages repeated at each time step: particle-to-grid (P2G) transfer, grid-based updates, and grid-to-particle (G2P) transfer. These operations are inherently parallel over particles and grid nodes, making them ideal targets for optimization.
JAX-MPM accelerates these stages using two key strategies: just-in-time (JIT) compilation to enable kernel fusion and vectorization to parallelize computation across simulation entities.

\textit{Remark 1 (JIT compilation for MPM kernels).}  
Each stage of the MPM update cycle—P2G, grid updates, and G2P—is wrapped using \lstinline{jax.jit}. This compiles Python functions into efficient XLA-executed code, enabling operation fusion and eliminating Python interpreter overhead. 

\textit{Remark 2 (Vectorized parallelism).}  
To exploit particle and grid parallelism, JAX-MPM combines \lstinline{jax.vmap} with vectorized array primitives. For example, polar decomposition is applied across particles with \lstinline{vmap}, while momentum transfer in P2G uses indexed scatter-add updates (\lstinline{x.at[idx].add(val)}) over all affected grid nodes. This removes explicit Python loops and enables efficient execution on both CPUs and GPUs.

\subsubsection{Memory optimization via checkpointing and segmentation}  

Reverse-mode AD incurs large memory costs by storing intermediates, especially in grid-based operations. JAX-MPM mitigates this using checkpointing and segmented backpropagation.

\begin{figure}[ht!]
    \centering
    \includegraphics[width=0.8\linewidth]{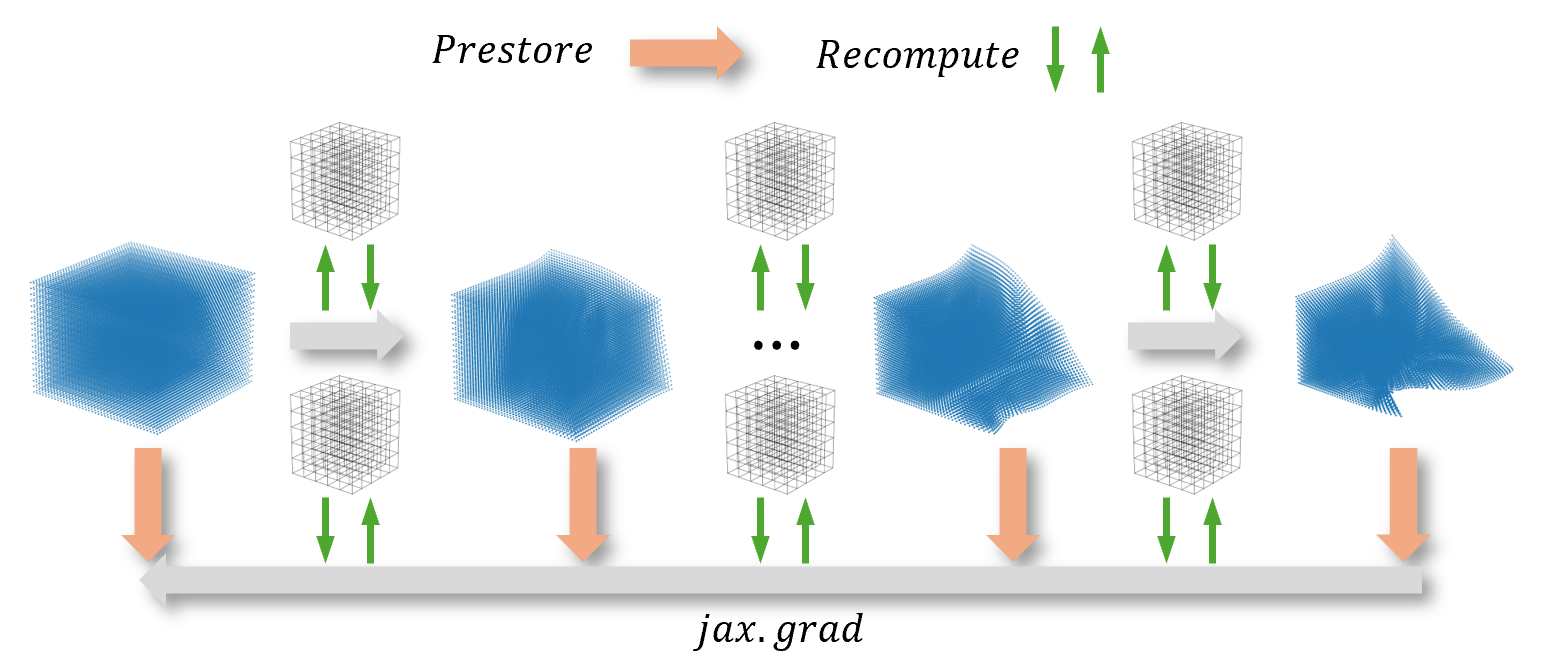}
    \caption{Backward propagation with checkpointed grid operations in JAX-MPM. 
During the forward pass, only particle states are stored while grid buffers are discarded.  During backpropagation, \lstinline{jax.checkpoint} recomputes the grid operations, reducing memory usage in reverse-mode automatic differentiation  (\lstinline{jax.grad}).}
    \label{fig:back}
\end{figure}

\textit{Remark 3 (Checkpointing grid operations).}
The particle–grid interactions, within MPM’s hybrid Eulerian-Lagrangian formulation, require substantial buffer storage during backpropagation.
To reduce memory usage, we apply gradient checkpointing: intermediate grid states are discarded during the forward pass and recomputed during the backward pass, while only particle-level outputs (e.g., $\bm{x}_p^t$, $\bm{v}_p^t$) are retained. Similar strategies have been used in prior work \cite{hu2019difftaichi, yuhn20234d}. 
Within the JAX-MPM framework, we implement this using \lstinline{jax.checkpoint} (also known as \lstinline{jax.remat}), which rematerializes intermediate computations in \lstinline{substep} during backpropagation, as illustrated in Fig. \ref{fig:back}. A representative implementation is shown below:

\begin{lstlisting}
@jax.remat
def substep(carry, _):
  v, x, rho, v_grad = carry
  base, fx, w, dw = pre_compute(x)
  rho, gv_in, gm_in = p2g(v, x, rho, v_grad, base, fx, w)
  gv_out = grid_op(gv_in, gm_in)
  v, x, v_grad = g2p(gv_out, v, x, base, fx, w, dw)
  return (v, x, rho, v_grad), None
\end{lstlisting}

\textit{Remark 4 (Segmented simulation loop with \lstinline{jax.scan}).}
To further reduce peak memory, the simulation loop is partitioned into \(n\) segments, each containing \(N_t/n\) time steps. Each segment is executed using \lstinline{jax.scan} and wrapped with \lstinline{jax.checkpoint}, allowing intermediate values within the segment to be discarded during the forward pass and recomputed during the backward pass. As a result, peak memory usage is determined by the buffer size needed to recompute a single segment, reducing memory complexity from \(\mathcal{O}(N_t)\) to \(\mathcal{O}(N_t/n)\), where \(N_t\) is the total number of time steps and \(N_t/n\) is the segment length.
An example implementation is shown below:
\begin{lstlisting}
@jax.checkpoint
def segment(carry):
  return lax.scan(step_fn, carry, None, length=block_size)

for i in range(num_segs):
  carry, _ = segment(carry)
\end{lstlisting}


%% file: results.tex
\section{Numerical results of predictive simulation}\label{sec:result}

This section begins with a validation of JAX-MPM on 2D benchmark problems, followed by large-scale 3D simulations that showcase its effectiveness in modeling complex phenomena such as free-surface flow, rigid-body contact, and granular dynamics, all efficiently accelerated on the GPU.

\subsection{Validation of JAX-MPM for 2D cases}

\subsubsection{Simulation of 2D dam-break problem}\label{sec:dam_analytical}  
\begin{figure}
    \centering
    \includegraphics[width=1.0\linewidth]{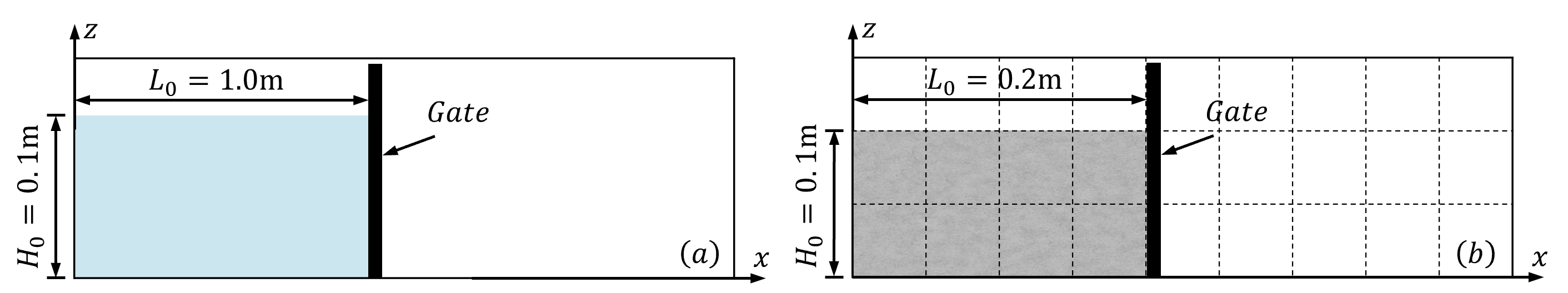}
    \caption{(a): Configuration of the 2D shallow dam-break problem~\cite{huang2015large}; (b): Schematics of the quasi-2D granular collapse simulation (corresponding to the experimental configuration of aluminum-bar assemblage~\cite{bui2008lagrangian}).}
    \label{fig:configure}
\end{figure}

Accurate simulation of free-surface dynamics is essential for geophysical applications such as floods, dam breaks, and debris surges.
To validate the capability of JAX-MPM in capturing such behavior, we simulate a classical two-dimensional dam-break problem under shallow-water assumptions (Fig.~\ref{fig:configure}a), following the setup in \cite{huang2015large}. The fluid is modeled as a weakly incompressible Newtonian fluid with zero viscosity and friction, enabling direct comparison with analytical results.

As illustrated in Fig.~\ref{fig:configure}a, a rectangular column of water with initial height \(H_0 = 100\,\mathrm{mm}\) and length \(L_0 = 1000\,\mathrm{mm}\) is released at \(t = 0\) by removing a gate on the right boundary. 
The simulation domain is discretized with a grid size of \(\Delta h = 0.004\,\mathrm{m}\), and each cell contains 4 particles. The numerical sound speed is \(c_n = 35\,\mathrm{m/s}\) and $\Delta t =1 \times 10^{-5}$ s is set to ensure numerical stability.
Slip boundary conditions are applied to all domain boundaries.

This benchmark admits an analytical solution
that characterizes the transient flow front propagation and surface evolution.
According to shallow-water theory \cite{stoker2019water, huang2015large}, the horizontal position of the  front at time \( t \) is given by
\begin{equation}
    x (t) = \left( 2 \sqrt{H_0 |g|} - 3 \sqrt{y |g|} \right)t + L_0,
    \label{eq:shallow}
\end{equation}
where \(g = -9.8\,\mathrm{m/s}^2\) is gravitational acceleration, and \(y\) is the vertical position at the flow front.

\begin{figure}
    \centering
    \includegraphics[width=1\linewidth]{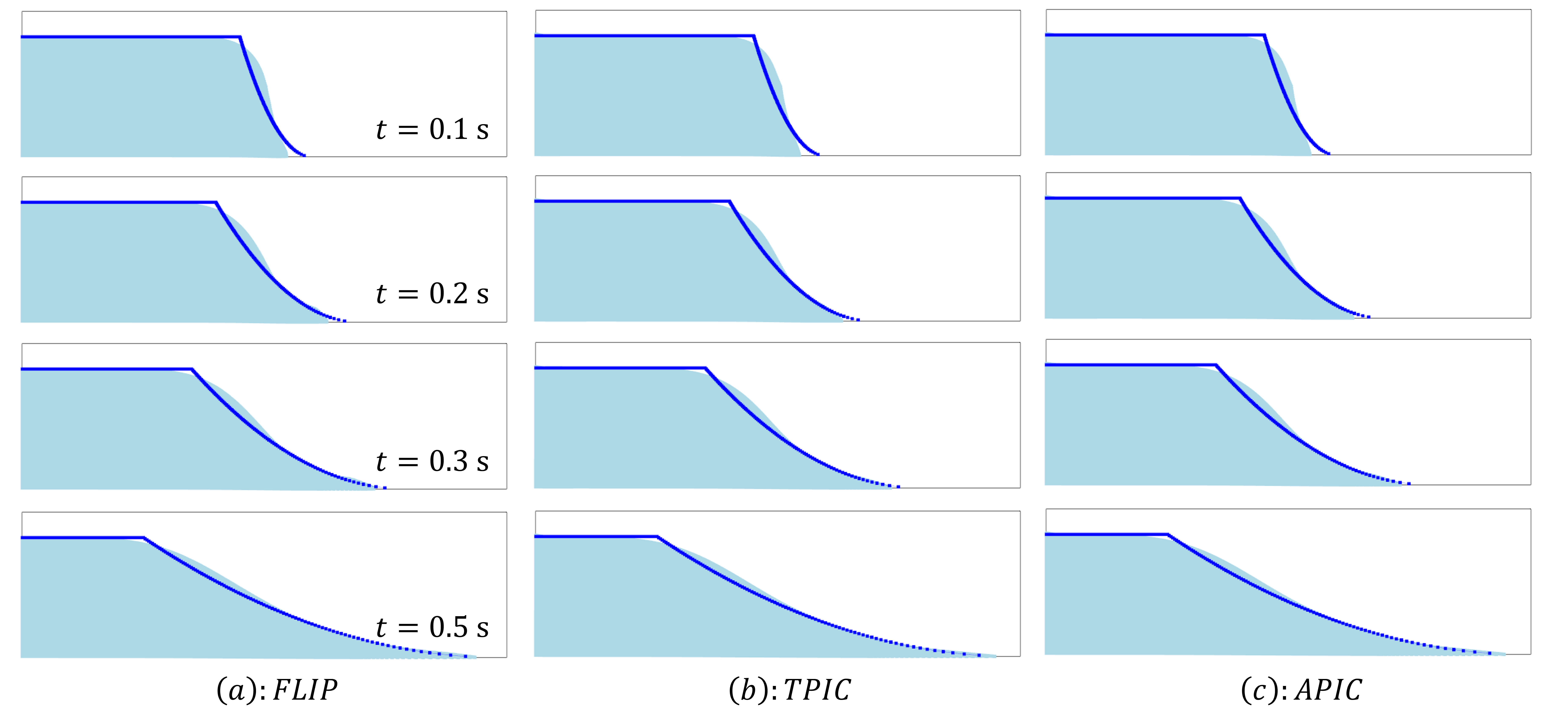}
    \caption{Simulation results of 2D shallow dam-break problem by JAX-MPM with (a) FLIP, (b) TPIC and (c) APIC.}
    \label{fig:2d_dam_break}
\end{figure}

Three  transfer schemes, including FLIP, TPIC, and APIC (see Section \ref{sec:g2p_trans}), are evaluated within the JAX-MPM framework for this benchmark. As shown in Fig.~\ref{fig:2d_dam_break}, all schemes produce results in close agreement with the analytical shallow-water solution. 
Minor  discrepancies observed during the early stage (\( t < 0.3\,\mathrm{s} \)) are primarily attributed to finite-depth effects and vertical accelerations that violate shallow-water assumptions. 
As the flow becomes shallower (\( t \geq 0.3\,\mathrm{s} \)), the numerical results exhibit improved agreement with theory. These observations are consistent with prior findings in \cite{huang2015large}, confirming the correctness of our implementation.

\subsubsection{Simulation of quasi-2D granular collapse: Aluminum-bar assemblage}\label{sec:granular}
As another 2D benchmark, we simulate the collapse of an aluminum-bar assemblage,
following  the experimental setup of Bui et al.~\cite{bui2008lagrangian}.
The material behavior is governed by the Drucker–Prager elastoplastic model, as described in Section \ref{sec:constitutive} and \ref{appendix:dp}.
The material is assumed to be non-cohesive ($c = 0$ kPa), with density \( \rho = 2650 \, \mathrm{kg/m}^3 \), bulk modulus \( K = 0.7 \, \mathrm{MPa} \), Poisson’s ratio \( \nu = 0.3 \), internal friction angle \( \varphi = 19.8^\circ \), and dilatation angle \( \psi = 0^\circ \). 

As shown in Fig.~\ref{fig:configure}b, the initial granular column has dimensions \( L_0 = 0.2 \, \text{m} \) and \( H_0 = 0.1 \, \text{m} \), and is initially confined by a vertical gate. 
To approximate a quasi-2D condition, the simulation is performed under a plane strain assumption within a computational domain \( (x, y, z) \in [0,\,0.6] \, \times [0,\,0.002] \, \times [0,\,0.12] \, \mathrm{m}^3 \), discretized into a uniform grid of \( 300 \times 1 \times 60 \) cubic cells. 
In the JAX-MPM simulation, no-slip boundary conditions are applied at the base, while slip conditions are imposed on all other boundaries. 
The FLIP transfer scheme is employed with a fixed time step \( \Delta t = 1 \times 10^{-5}\) s over a total simulated duration of \( 0.65 \, \mathrm{s} \).

\begin{figure}
    \centering
    \includegraphics[width=1\linewidth]{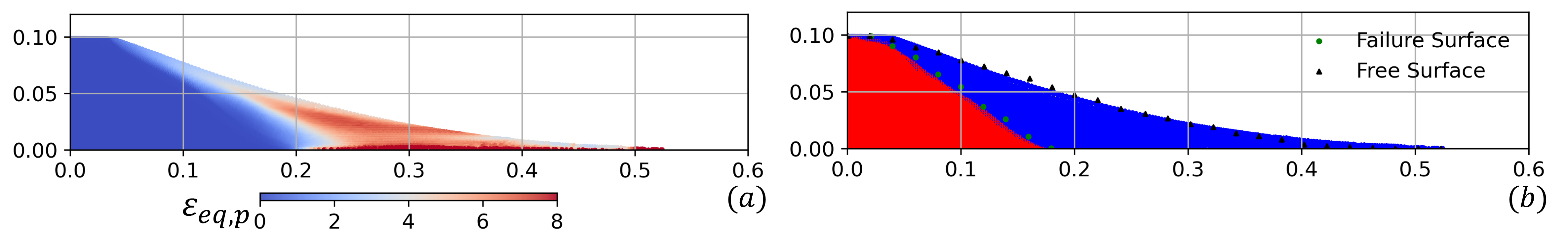}
\caption{Simulation results of the aluminum-bar assemblage collapse using JAX-MPM. 
(a) Contour plot of equivalent plastic strain \( \varepsilon_{eq,p} \), illustrating shear band development. 
(b) Comparison with experimental data from Bui et al.~\cite{bui2008lagrangian}, where 
the green dots and black triangles denote the experimentally observed failure surface and free surface, respectively. The simulated failure surface is defined by a threshold \( \varepsilon_{eq,p} > 0.03 \).}
    \label{fig:2d_flip}
\end{figure}

Figure~\ref{fig:2d_flip}a shows the contour of the simulated equivalent plastic strain \( \varepsilon_{eq,p} \), with peak values concentrated near the base and propagating upward to form localized shear bands.
This is consistent with typical shear localization behavior observed in granular collapse
(cf. Fig. 15 in \cite{nubel2004study}). 
 Fig.~\ref{fig:2d_flip}b presents the predicted free surface and failure surface, defined by a threshold of \( \varepsilon_{eq,p} > 0.03 \), both showing good agreement with experimental observations in~\cite{bui2008lagrangian}.
 
These results demonstrate that the proposed JAX-MPM framework with the Drucker–Prager model accurately captures  key features of granular flow behavior, including shear localization and free-surface evolution. 
Moreover, this benchmark highlights 
the robustness of our framework in modeling elastoplastic deformation and large-deformation dynamics in 
geological materials.

\subsection{Application of JAX-MPM for 3D cases}
To evaluate the effectiveness of JAX-MPM in modeling complex 3D geophysical dynamics, we consider three benchmark scenarios: (i) a 3D dam-break with embedded rigid obstacles 
(Section \ref{sec:3D_dam_break}), (ii) 3D granular column collapse (Section~\ref{sec:forward_column}), and (iii) 3D granular cylinder collapse (Section~\ref{sec:forward_cylinder}). 
The corresponding simulation configurations and computational costs are summarized in Table~\ref{tab:simulation_params}.

\subsubsection{Simulation of 3D dam-break problem}\label{sec:3D_dam_break}

\begin{figure}
    \centering
    \includegraphics[width=0.65\linewidth]{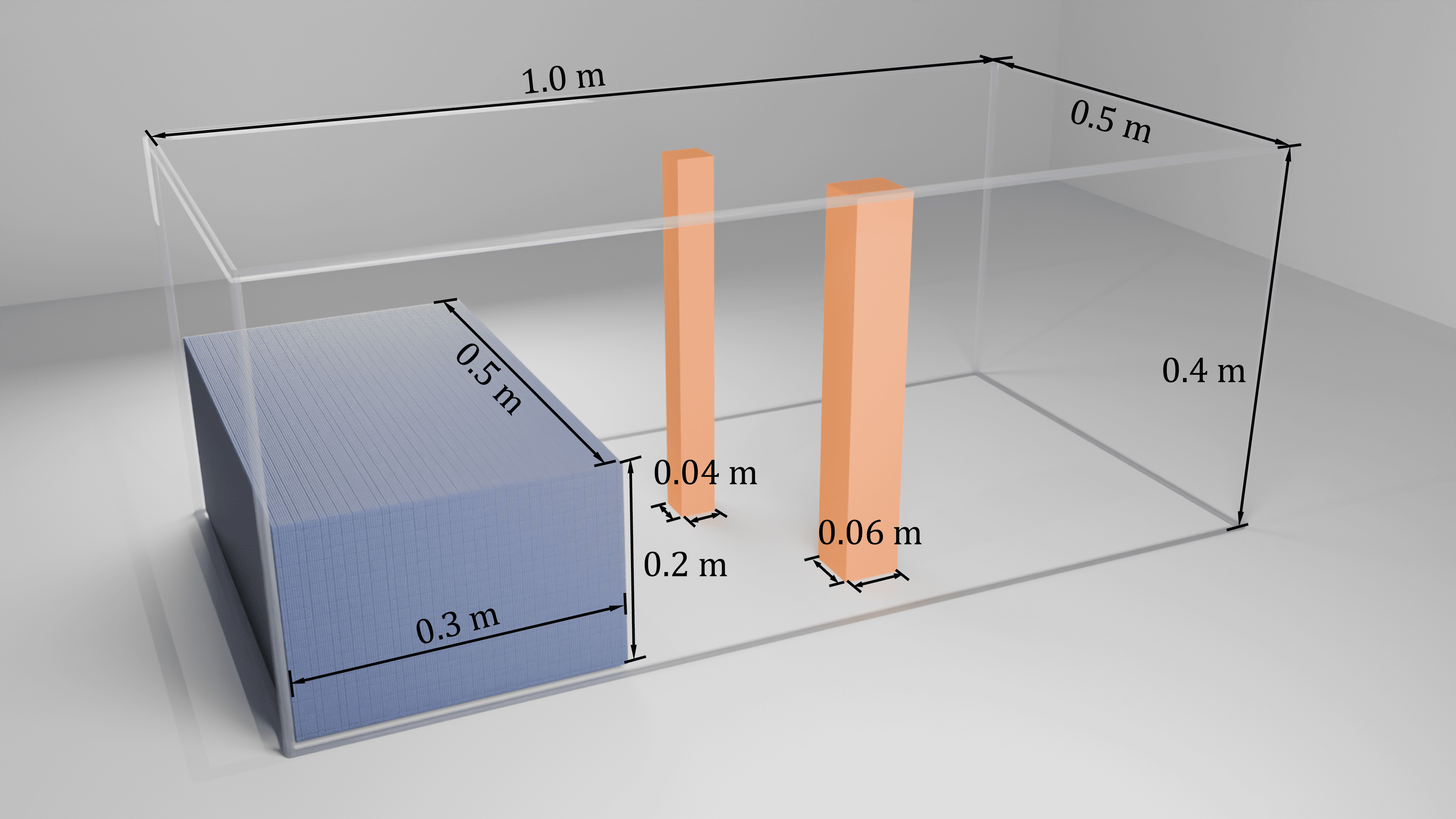}
    \caption{Initial configuration of 3D dam-break simulation with two rigid internal obstacles.}
    \label{fig:config_3d}
\end{figure}

As illustrated in Fig.~\ref{fig:config_3d}, the computational domain for the 3D dam-break scenario  is defined as 
\( [0,\,1.0] \times [0,\,0.5] \times [0,\,0.4] \, \mathrm{m}^3 \).
The initial dimensions of the water column are \( H_0 = 0.2 \, \mathrm{m} \) (height), \( L_0 = 0.3 \, \mathrm{m} \) (length), and \( W_0 = 0.5 \, \mathrm{m} \) (width). The fluid is modeled as a weakly compressible Newtonian fluid with a dynamic viscosity of \( \mu = 1.01 \times 10^{-3} \, \mathrm{Pa \cdot s} \) and a numerical sound speed of \( c_n = 35 \, \mathrm{m/s} \). 

Two rigid square columns are placed in the domain as internal obstacles. Their side lengths are 0.04 m and 0.06 m, positioned at \( (0.5, 0.3, 0.0) \, \mathrm{m} \) and \( (0.6, 0.1, 0.0) \, \mathrm{m} \), respectively. 
A frictionless contact model employing a \textit{predictor–corrector scheme} \cite{bardenhagen2000material} is used to prevent fluid penetration.
Since the rigid obstacles are stationary, contact enforcement reduces to nullifying the fluid's normal velocity component at the interface:
\begin{equation}
\bm{v}_i = \tilde{\bm{v}}_i - \left(\tilde{\bm{v}}_i \cdot \bm{n}_i \right) \bm{n}_i,
\end{equation}
where \( \tilde{\bm{v}}_i \) is the predicted fluid velocity vector (\textit{predictor}),  \( \bm{v}_i \) is the corrected velocity vector, and \( \bm{n}_i \) is the normal vector to the rigid surface. 
This correction allows slip along the boundaries by preserving the tangential velocity component.

\begin{figure}
    \centering
    \includegraphics[width=1.0\linewidth]{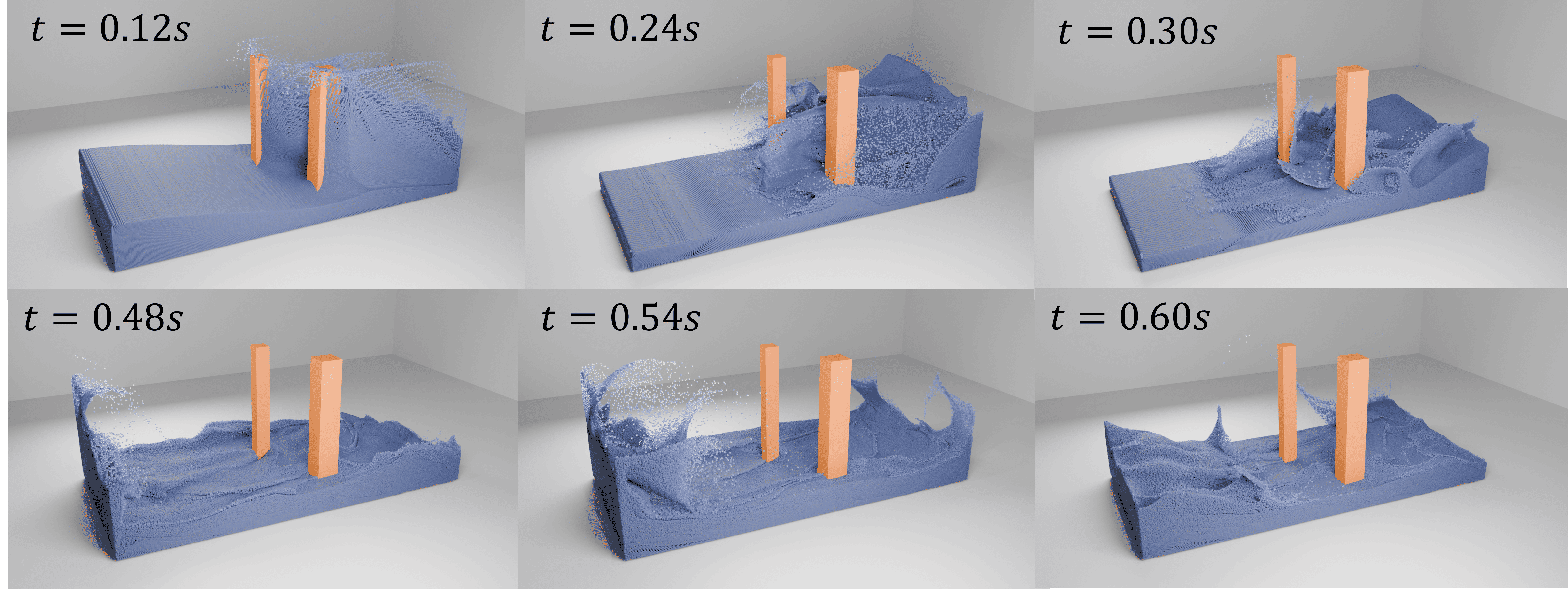}
\caption{Snapshots of 3D dam-break simulation with rigid internal obstacles. 
The sequence from $t=0.12$ to $0.60$ s captures the fluid front impacting the columns, generating upward splashing, and subsequently forming recirculation zones and lateral flow separation downstream.}
    \label{fig:contact_dam}
\end{figure}

Fig.~\ref{fig:contact_dam} illustrates  the dynamic evolution of the free-surface flow interacting with embedded rigid obstacles. 
During the initial stages (\( t = 0.12 \)–\( 0.30\,\mathrm{s} \)), the fluid front impacts the columns, producing upward splashing and significant surface deformation. 
At later times, recirculation zones and lateral flow separation emerge downstream of the obstacles,  
demonstrating the framework’s ability to capture complex three-dimensional fluid behaviors induced by embedded structures.

Besides capturing physically realistic flow dynamics, JAX-MPM showcases excellent computational efficiency.
The simulation presented in Fig.~\ref{fig:contact_dam}, featuring over 1.9 million particles and a \( 200 \times 100 \times 80 \) grid. It runs on GPU with JIT compilation and completes 1000 time steps in approximately 8 seconds (see Table~\ref{tab:simulation_params}). 
This computational performance underscores the scalability and efficiency of the JAX-based implementation. 

\begin{table}[h]
\centering
\small
\caption{Summary of simulation configurations and computational costs.
The label ``sim'' refers to forward simulations, with runtimes reported for both single precision (float32) and double precision (float64), formatted as ``float32 / float64''. 
The label ``sim + grad'' denotes inverse simulations involving both forward and gradient computations, performed exclusively in double precision (float64). ``s/1000 steps'' indicates the wall time (seconds) for every 1000 time steps.}
\scriptsize
\begin{tabular}{l|c c c c c}
\hline
Example & Particle \# & Grid resolution & Param \# & s/1000 steps & $\Delta h$ \\
\hline
Dam-break (Fig. \ref{fig:contact_dam}) & 1,920,000& $200 \times 100 \times 80$ & N/A & 8.2/35.1 (sim)& $1 / 200$ \\
Granular column (Fig. \ref{fig:3d_render_granular}) & 2,408,448 & $400 \times 32 \times 200$ & N/A & 16.2/69.3 (sim)& $1 / 400$  \\
Granular cylinder (Fig. \ref{fig:3d_render_granular_cylinder}) & 2,712,960 & $400 \times 400 \times 100$ & N/A & 22.3/98.4 (sim)& $1 / 400$ \\
\multirow{2}{*}{Inv. velocity (Fig. \ref{fig:init_profile})} 
  & 10,000 & $150 \times 60$ & 1951 & 0.62 (sim + grad) & $1 / 100$ \\
  & 40,000 & $300 \times 120$ & 1951 & 1.52 (sim + grad) & $1 / 200$ \\
Inv. friction (Fig. \ref{fig:fric_multiple}) & 6000 & $200 \times 40$ & 5 & 0.59 (sim + grad)& $1 / 100$ \\
\hline
\end{tabular}
\label{tab:simulation_params}
\end{table}

\subsubsection{3D granular column collapse: Influence of aspect ratio}\label{sec:forward_column}

To evaluate the capability of JAX-MPM in simulating large-scale granular flow behaviors and capturing the influence of geometric variation, we consider the 3D granular column with varying initial aspect ratios.
The objective is to investigate how the geometry, characterized by the aspect ratio  $a = H_0/L_0$, affects the flow dynamics, runout distance, and failure mechanisms. 
Here, 
$H_0$ and $L_0$ denote the initial height and base length of the granular column, respectively. 
Simulations are carried out for
a total of 200{,}000 time steps, corresponding to a physical duration of 2\,s with a time step size of \( \Delta t = 10^{-5} \,\text{s} \).
The material properties remain the same as in Section \ref{sec:granular}. 

\begin{figure}[htb]
    \centering
    \includegraphics[width=1\linewidth]{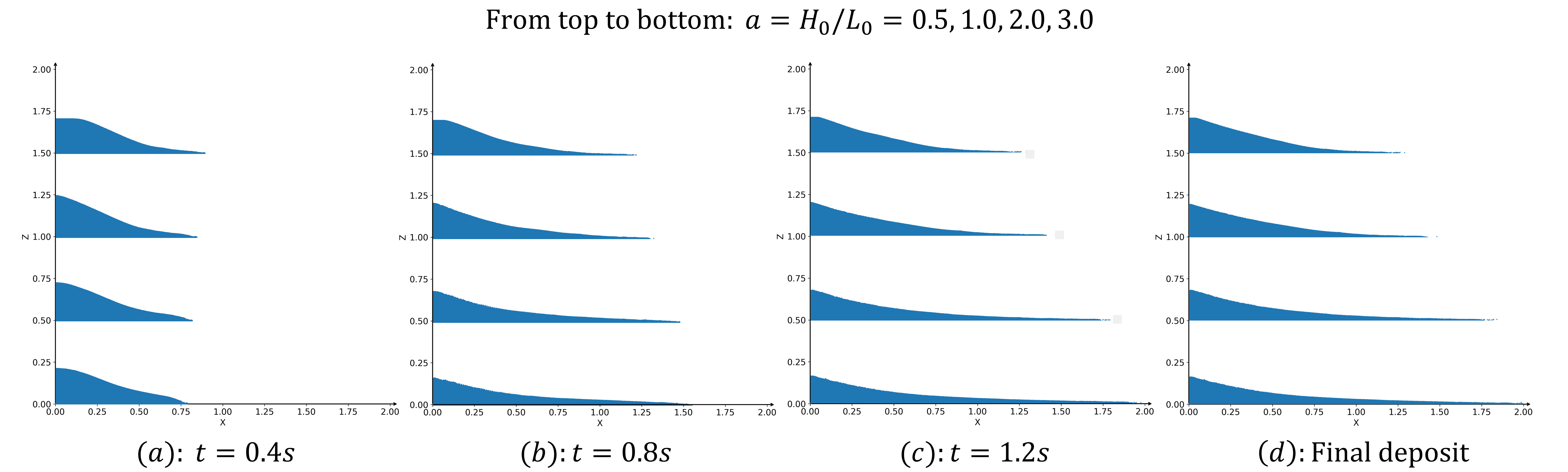}
    \caption{Simulation results of 3D granular column collapse with varying aspect ratios. 
    2D deposit profiles are shown at \( t = 0.4\,\text{s},\, 0.8\,\text{s},\, 1.2\,\text{s},\, \text{and}\,\text{final deposit} \) with initial aspect ratios \( a = H_0/L_0 = 0.5, 1.0, 2.0, 3.0 \). Higher aspect ratios lead to greater vertical collapse and longer runout distances due to enhanced free-fall and mass mobilization.}
    \label{fig:3d_render_granular}
\end{figure}

Four aspect ratios are investigated: 
\( a = 0.5, 1.0, 2.0, \) and \( 3.0 \). 
As shown in Fig.~\ref{fig:3d_render_granular},
increasing the aspect ratio results in a larger portion of the column undergoing free fall, leading to reduced static basal zones and enhanced lateral spreading.

Moreover, the final runout distances \( L_f \) are compared by defining the normalized runout as \( d_n = (L_f - L_0)/L_0 \).
The results show a clear increase in $d_n$ with  aspect ratio:
2.05 (\(a = 0.5\)), 3.97 (\(a = 1.0\)), 7.76 (\(a = 2.0\)), and 10.65 (\(a = 3.0\)).
The trend transitions from linear to sublinear scaling, consistent with experimental observations by Staron and Hinch~\cite{staron2005study}, where \( d_n \propto a \) for small \( a \) and \( d_n \propto a^{0.7} \) for larger values.

\subsubsection*{Computational scalability}\label{sec:forward_column_time}
To assess the computational performance, we benchmark JAX-MPM against the open-source C++ solver {CB-Geo}~\cite{kumar2019scalable}. 
Five discretization resolutions are tested, with grid spacings ranging from \(\Delta h = 0.015\) to \(\Delta h = 0.004\), corresponding to particle counts from approximately \(1.7 \times 10^4\) to \(9.0 \times 10^5\). 
Wall-clock times are recorded for every 1000 simulation steps under the following hardware configurations:
(1) CB-Geo on an AMD EPYC 7763 CPU with 128 threads, and (2) JAX-MPM on an NVIDIA A100 GPU.
Note that CB-Geo employs a Mohr–Coulomb material model, as the built-in solver does not currently support Drucker–Prager plasticity. 
While Mohr–Coulomb models are generally more expensive due to their non-smooth yield surfaces, the computational complexity is comparable, and thus the comparison still provides meaningful insight into relative scalability.

\begin{figure}[htb]
    \centering
    \includegraphics[width=0.6\linewidth]{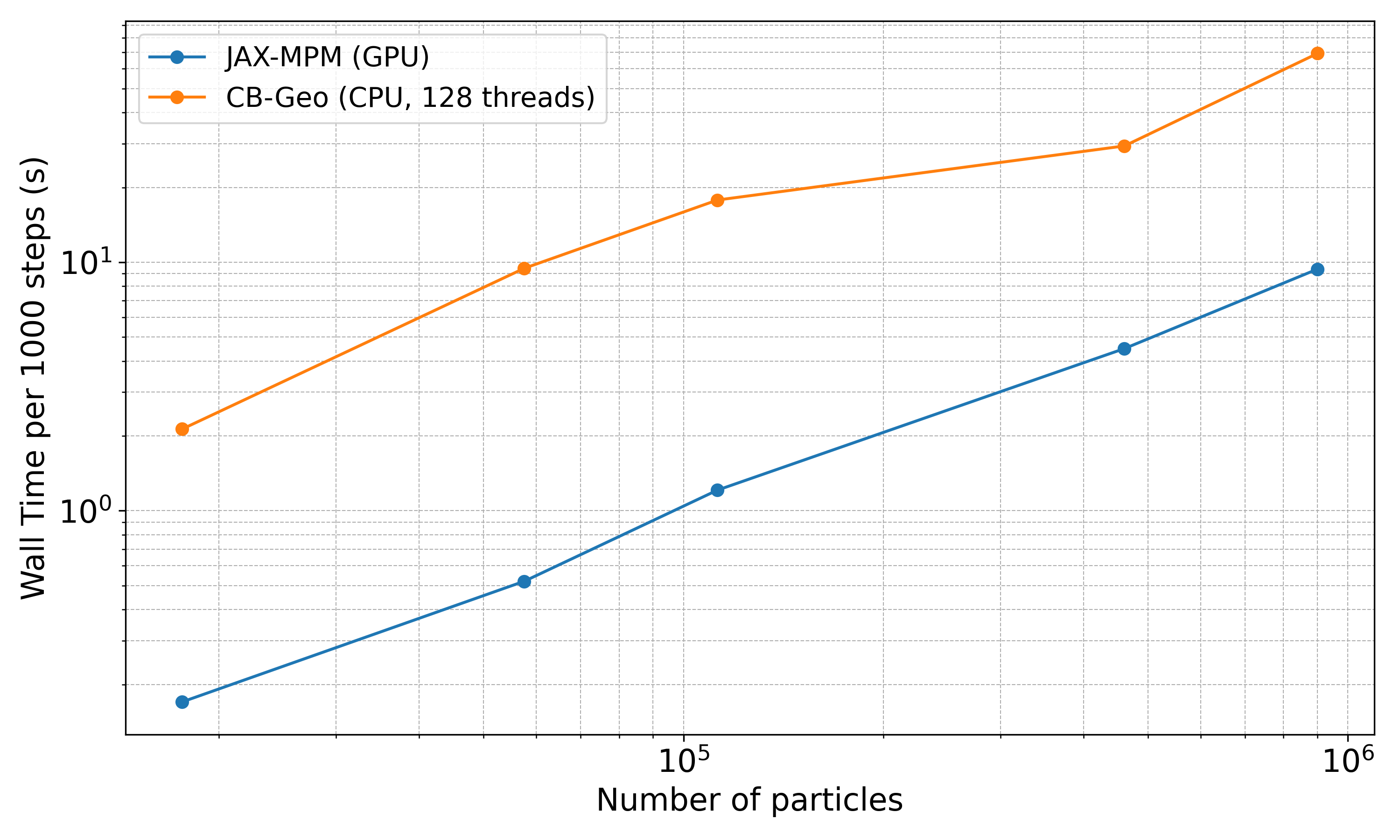}
    \caption{Computational scaling (wall-clock time)  
    comparison between JAX-MPM (GPU) and the open-source C++ solver CB-Geo (CPU) ~\cite{kumar2019scalable} across five simulation resolutions (from approximately 17.6k to 900k particles). All simulations are performed in double precision (float64).
    } 
    \label{fig:efficiency_comparison}
\end{figure}

As shown in Fig.~\ref{fig:efficiency_comparison}, JAX-MPM executed on a GPU consistently outperforms 
CB-Geo on CPU configurations across all tested resolutions.
At the highest resolution (\(\Delta h = 0.004\), 900,000 particles), 
JAX-MPM completes 1000 steps in roughly 9 seconds,
achieving more than a 7\(\times\) speedup compared to CB-Geo.

\subsubsection{3D granular cylinder collapse: Influence of friction angle}\label{sec:forward_cylinder}

We further assess the capability of JAX-MPM by simulating the collapse of a 3D granular cylinder with varying internal friction angles: \( \varphi = 20^\circ, 25^\circ, 35^\circ, \) and \( 40^\circ \).
Each simulation runs for 100{,}000 steps (1\,s of physical time) with a time step size of \( \Delta t = 10^{-5} \,\text{s} \).
The Drucker–Prager elastoplastic model is employed, with all other material parameters remaining identical to those detailed in Section \ref{sec:granular}.

\begin{figure}
    \centering
    \includegraphics[width=1\linewidth]{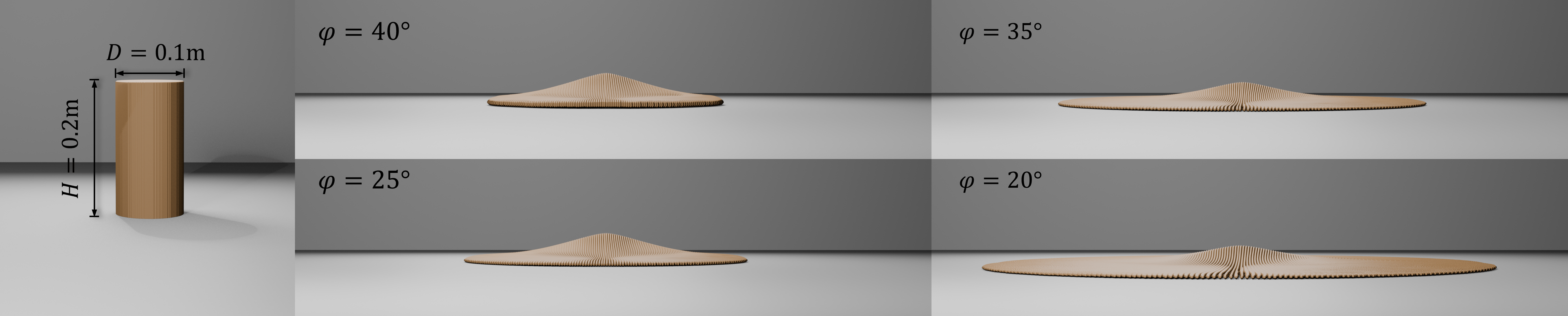}
    \caption{3D granular cylinder collapse with varying friction angles.
    Final deposit profiles for friction angles \( \varphi = 20^\circ, 25^\circ, 35^\circ, 40^\circ \). Higher friction results in steeper, more confined piles and reduced spreading due to increased shear resistance.}
    \label{fig:3d_render_granular_cylinder}
\end{figure}

Fig.~\ref{fig:3d_render_granular_cylinder} shows that increasing the friction angle produces steeper, more confined deposits and shorter runout, indicating greater resistance to shear deformation. These trends are consistent with theoretical and experimental studies reported in the literature~\cite{kumar2023experimental, klar2016drucker} and highlight the friction angle as a primary control on shear resistance, flow mobility, and final deposit geometry.

\subsubsection*{Computational performance discussion}\label{sec:forward_cylinder_time}
The JAX-MPM simulation involves over 2.7 million particles on a \( 400 \times 400 \times 100 \) computational grid  (approximately 16 million grid nodes). On an NVIDIA A100 GPU, JAX-MPM completes 1000 simulation steps in approximately 22 seconds (in single precision) and 98 seconds (in double precision). 
For comparison, 
a recent high-performance Fortran-based MPM solver with dynamic load balancing (DLB) reported a runtime of 688 seconds for a similar scenario on 288 CPU cores, 
with a comparable material model and a grid resolution of 
\( 280 \times 280 \times 120 \) (approximately 9.4 million grid nodes)~\cite{hidano2025b}. 
Without DLB, the runtime increases significantly to roughly 13,803 seconds.
These comparisons show that JAX-MPM achieves a \( 7\times \)  speedup over the optimized Fortran+DLB implementation and over \( 140\times \) speedup relative to the baseline CPU version without DLB.

Together with the benchmarking results in Section~\ref{sec:forward_column_time}, these comparisons highlight the computational efficiency and scalability of JAX-MPM. The framework’s performance is enabled by just-in-time (JIT) compilation, GPU parallelism, and kernel fusion (see Section~\ref{sec:speedup}), making it well suited for large-scale geomechanical simulations. 

In the following section, 
we further demonstrate the unique capability of JAX-MPM to perform inverse modeling through end-to-end differentiation, a task that is often cumbersome or intractable using conventional solvers.

\section{Inverse modeling of geophysical flows using JAX-MPM}\label{sec:Inverse_dam}

Inverse modeling is essential in geophysical flow analysis, as it enables the recovery of unobservable system parameters and supports the development of effective mitigation strategies. Parameters such as initial velocity fields and spatially varying basal friction~\cite{he2023hybrid} fundamentally influence runout extent, flow mobility, and energy dissipation, yet they are often inaccessible through direct measurement. 
In these contexts, inverse modeling becomes a key tool for post-event analysis, model calibration, and data-informed forecasting.

Prior studies have approached inverse problems in geomechanics using techniques  such as traditional back-analysis~\cite{calvello2017role}, Bayesian inference with surrogate models~\cite{zhao2022bayesian}, and neural network simulators~\cite{choi2024inverse,choi2025gns}. 
More recently, there has been growing interest in leveraging gradient-based optimization and algorithmic automatic differentiation to enable efficient parameter inference and process learning in geoscience and solid mechanics applications ~\cite{gelbrecht2023differentiable,shen2023differentiable,du2024differentiable}.

In this section, we further evaluate the JAX-MPM framework’s capability for learning-based inverse modeling through a series of parameter-estimation tasks that are challenging for traditional non-differentiable solvers.
Two representative problems in a dam-break setting are considered: 
(1) recovery of unknown initial velocity fields (Sections~\ref{sec:inverse_velocity}–\ref{sec:inverse_velocity_nn}) and (2) estimation of spatially variable friction coefficients (Section~\ref{sec:inverse_friction}) from sparse dynamic measurements.
Unless otherwise specified, all inverse modeling experiments in this section are optimized using the Adam algorithm with a decaying learning rate starting at \( lr = 0.1 \).

\subsection{Inverse estimation of initial velocity constant}\label{sec:inverse_velocity}
We first test the inverse estimation of a scalar parameter that defines the  initial horizontal velocity field in a 2D dam-break scenario (see Section \ref{sec:dam_analytical}). 
The ground-truth initial velocity field is prescribed as  \( v_x^0 = \overline{\alpha}(H_0 - y) \) and \( v_y^0 = 0 \), with  \( \overline{\alpha} = 2.0 \). For the inverse modeling task,
\( \overline{\alpha}\) is treated as an unknown coefficient to be inferred.

The reference simulation model assumes an inviscid fluid (\( \mu = 0 \)) with a numerical sound speed of \( c_0 = 50 \) m/s.
The initial water column has height and length \( H_0 = L_0 = 0.5 \) m, 
 discretized using a grid spacing of \(\Delta h = 0.01\,\mathrm{m}\) with four particles per cell. 
The simulation is run for a total duration of 
\( t = 0.3 \) s with a fixed time step \( \Delta t = 3 \times 10^{-5} \) s.

The inverse problem is formulated as a PDE-constrained optimization~\eqref{eq:inverse_loss}, where
the objective is to minimize the discrepancy between simulated dynamics and reference observations over the final
\( |\mathcal{T}_t| = 100 \) frames,
sampled every 10 steps from the last 1000 simulation steps (i.e., the final 0.03 s of the simulation).

We consider three observation strategies, each characterized by a specific choice of simulation variable \( \boldsymbol{z}^t \) and spatial observation indices \( \mathcal{T}_x\):

\textbf{Case 1} - Full particle position supervision (Lagrangian): All material points are tracked, with the observation variable defined as particle positions, i.e., \( \boldsymbol{z}^t = \boldsymbol{x}^t \). 
The observed quantity is then  \( \mathcal{Q}_l(\boldsymbol{z}^t) = \boldsymbol{x}_l^t \) for each index \( l \in \mathcal{T}_x \), with \( |\mathcal{T}_x|= \) 10,000.

\textbf{Case 2} - Sparse particle position supervision (Lagrangian): A randomly selected subset of \( |\mathcal{T}_x| = 100 \) particles is tracked, with the same observation mapping as in the full supervision Case 1.

\textbf{Case 3} -  Sparse velocity-monitor supervision (Eulerian): 
Predefined spatial regions are used to monitor average particle velocities.  The simulation variable is set to the particle velocities,  \( \boldsymbol{z}^t = \boldsymbol{v}^t \), and the observed simulated quantity is computed as:
\begin{equation}
\mathcal{Q}_l(\boldsymbol{z}^t) = \frac{1}{|\mathcal{P}_l^t|} \sum_{p \in \mathcal{P}_l^t} \boldsymbol{v}_p^t, \quad  \mathcal{P}_l^t = \{ p \mid \boldsymbol{x}_p^t \in \Omega_l\}
\end{equation}
where \( \mathcal{P}_l^t\) denotes the set of particles located within the $l$-th Eulerian monitor region \( \Omega_l \). 
For example, as illustrated in 
Fig.~\ref{fig:geometry}, we consider \( N_m = 9 \) monitor regions, each region is a square with a half-side length of 0.01\,m.
The centers of these regions are arranged in
a \( 3 \times 3 \) grid
at spatial locations \(\{0.2, 0.3, 0.4\} \text{ m} \times \{0.1, 0.2, 0.3\} \text{ m}\).

\begin{figure}[htb]
    \centering
    \includegraphics[width=1.0\linewidth]{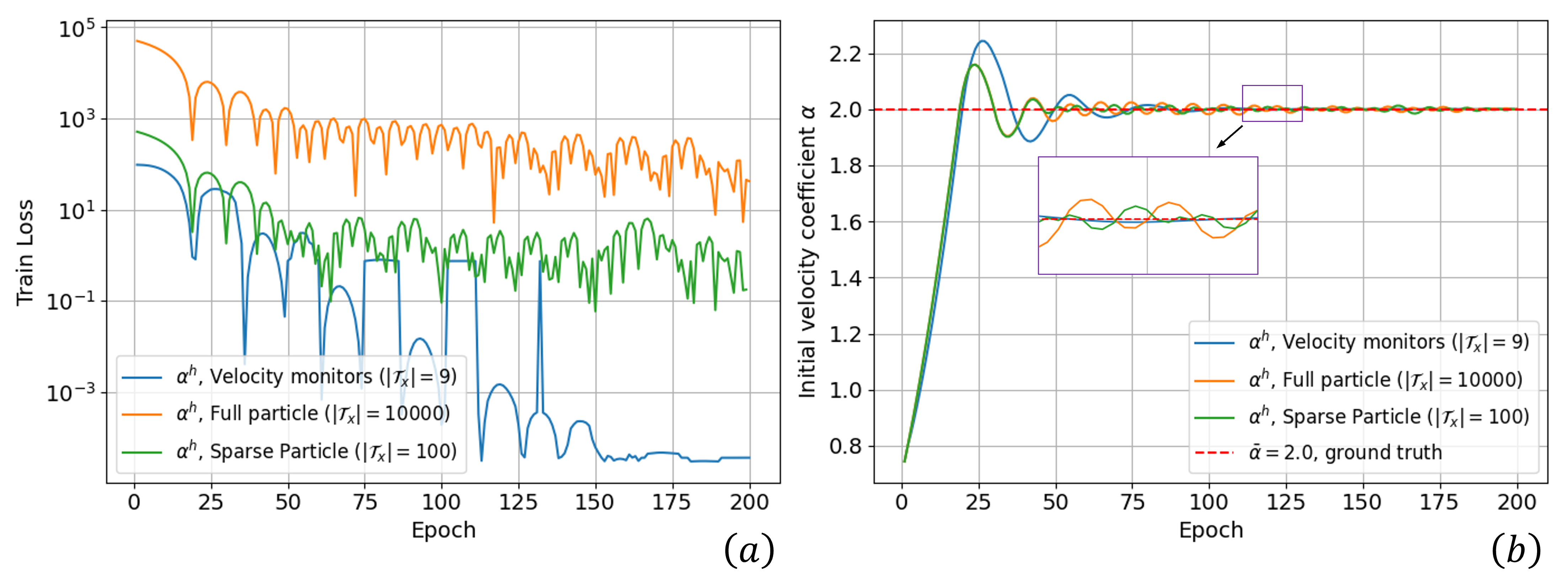}
    \caption{Training evolution for inverse recovery of the initial velocity constant \( \alpha \) under three supervision strategies. (a): training loss over epochs. (b): evolution of the estimated parameter \( \alpha^h \).}

    \label{fig:loss_break}
\end{figure}

\begin{figure}[htb]
    \centering
    \includegraphics[width=1\linewidth]{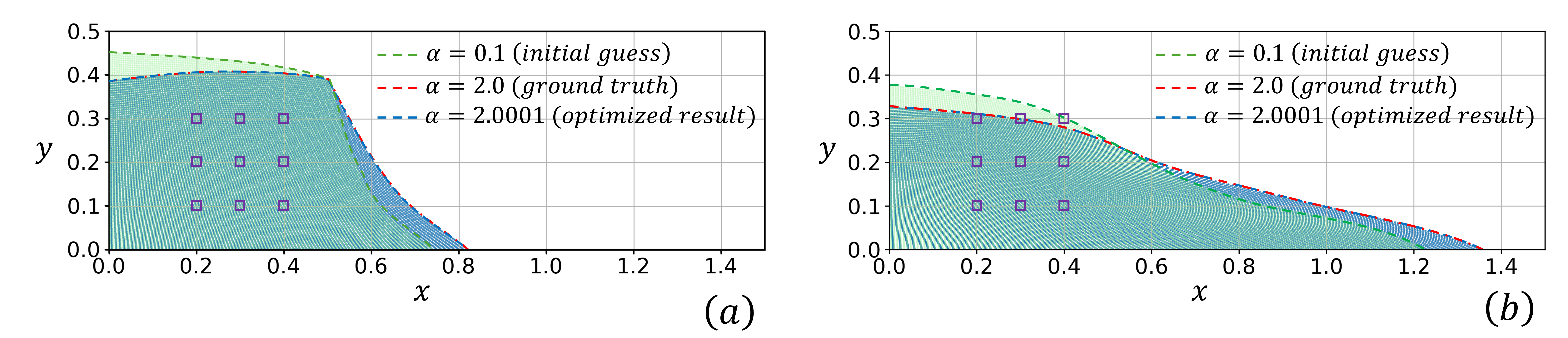}
    \caption{Comparison of flow geometries at (a) $t = 0.15$ s and (b) $t = 0.30$ s for the initial guess, ground truth, and optimized initial velocity, obtained by using the velocity monitor-based (Eulerian) supervision. 
    Purple squares indicate the locations of the $N_m = 9$ velocity monitors.
    }
    \label{fig:geometry}
\end{figure}

The JAX-MPM  optimization begins from an initial guess \( \alpha_0 = 0.1 \).
Fig.~\ref{fig:loss_break} presents 
the evolution of both the training loss and the estimated parameter 
\( \alpha^h \)
over the course of learning iterations, demonstrating that all three observation strategies successfully converge to the 
ground truth (\( \overline{\alpha} = 2.0 \)). 
Among them, velocity-monitor supervision achieves the fastest and most stable convergence, despite using the fewest observations.
This reflects a strong alignment between the supervised quantity and the underlying parameter, as 
\( \alpha \) directly governs the velocity field.
Notably, the sparse particle supervision (Case 2) achieves convergence comparable to—or even faster than—that of full particle supervision (Case 1). 
We conjecture that, while full-field observations provide denser information, they may also introduce greater complexity into the loss landscape and its gradients, resulting in  increased oscillations during optimization, particularly when the underlying parameter is spatially uniform.

Fig.~\ref{fig:geometry} visualizes the flow fields at \( t = 0.15 \) s and \( t = 0.30 \) s for the initial guess, the ground truth, and the optimized result obtained using the velocity monitor-based (Eulerian) supervision. The final deposit shape closely matches the reference simulation, indicating that the inferred parameter leads to physically accurate flow dynamics. These results confirm the effectiveness of JAX-MPM in solving inverse problems using both Lagrangian and Eulerian observation types within a unified, fully end-to-end differentiable framework.

\subsection{Inverse estimation of initial velocity field}\label{sec:inverse_velocity_nn}
We extend the previous example to a more general scenario in which the initial horizontal velocity field varies spatially and cannot be represented by a single scalar parameter.
The reference horizontal velocity field is defined as
\begin{equation}
v_x^0(y) = 2.0 \left(1 - \left( \frac{y}{H_0} \right)^2 \right) + 0.2 \sin\left(4\pi \frac{y}{H_0}\right),
\end{equation}
which combines a parabolic base profile with a sinusoidal perturbation.

The objective is to recover the continuous velocity field \( v_x^0(y) \) from the observational data. 
To this end, we parametrize the field using a feedforward neural network \( \mathcal{N}_\theta \) (c.f. Eq. \ref{eq:nn_apprx}), consisting of three hidden layers with 30 ReLU-activated units  each. 
The network maps the vertical coordinate \( y \) to the corresponding velocity \( v_x^0(y) \), where \( \theta \) denotes the trainable weights.

Since the goal is to reconstruct a continuous spatial velocity field, velocity monitor data are excluded in this experiment to better showcase the applicability of the proposed framework. 
Instead, we adopt the sparse particle (Lagrangian) position  supervision scheme introduced in Section~\ref{sec:inverse_velocity}. To assess the effect of observation sparsity, we randomly sample particles  at three resolutions: 100, 500, and 1000 particles (\( |\mathcal{T}_x| \in \{100, 500, 1000\} \)).
The trajectories of these sampled particles constitute the observation dataset.
In all cases, the loss is evaluated over the final \( |\mathcal{T}_t| = 100 \) frames (last 0.03\,s) sampled from the last 1000 time steps, which is consistent with the experiment setup in Section \ref{sec:inverse_velocity}. 

\begin{figure}
    \centering
    \includegraphics[width=1\linewidth]{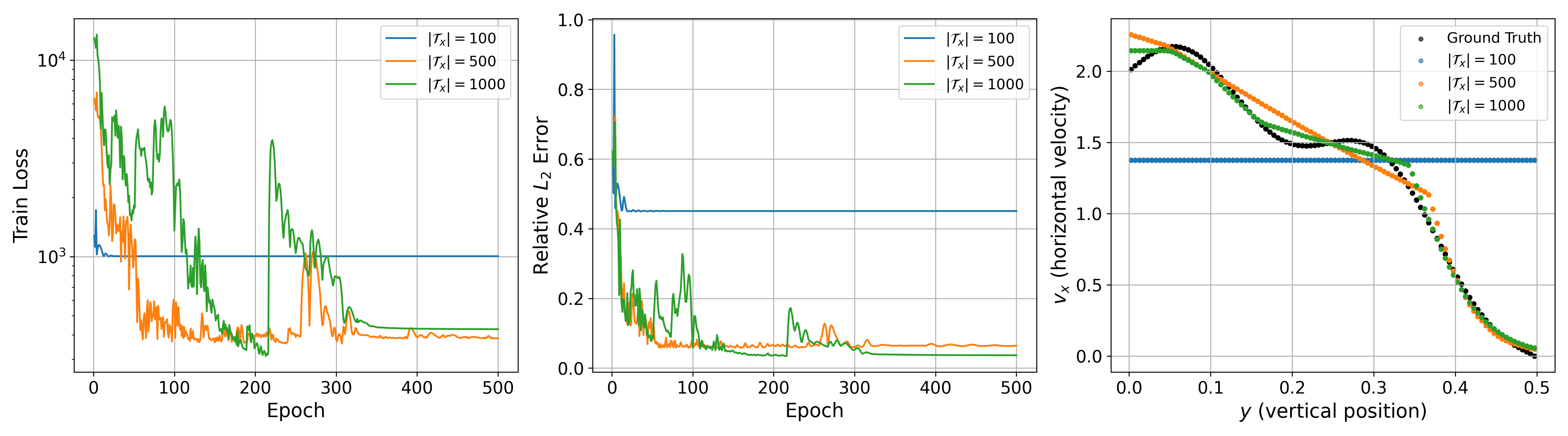}
    \caption{Inverse recovery of a spatially varying initial velocity field using a neural network. (a) Training loss over epochs, (b) convergence of relative $L_2$ error, and (c) comparison between the learned and reference velocity profiles.}
    \label{fig:init_profile}
\end{figure}

As shown in Fig.~\ref{fig:init_profile}(a–b), the training loss for \( |\mathcal{T}_x| = 500 \) and \(|\mathcal{T}_x| = 1000 \) decreases steadily and converges after around 400 epochs. In contrast, the loss for \( |\mathcal{T}_x| = 100 \) remains nearly flat throughout training, indicating that the 
observation density is insufficient to drive meaningful parameter updates.
The final relative \( L_2 \) errors between predicted and reference velocity fields are 45.1\% for 100 observation particles, 6.4\% for 500, and 3.7\% for 1000. 
These results suggest there could exist a threshold in observation density for effective field reconstruction in this setting.

Fig.~\ref{fig:init_profile}c compares the reconstructed initial velocity profiles against the ground truth. Both the \( |\mathcal{T}_x| = 500 \) and \( |\mathcal{T}_x| = 1000 \) cases accurately capture the overall profile, especially in the upper region. 
The \( |\mathcal{T}_x| = 1000 \) case provides a smoother and more accurate reconstruction, 
while the \( |\mathcal{T}_x| = 500 \) case slightly underestimates local variations due to limited spatial coverage, which restricts the model’s ability to resolve finer features.
In contrast, the \( |\mathcal{T}_x| = 100 \) case fails to recover any meaningful spatial structure, resulting in a nearly constant initial velocity prediction. 
This again underscores  
the importance of sufficient observation density for successful continuous field reconstruction.

Deviations near the lower boundary (i.e., close to $y=0$) reveal a common challenge in inverse field recovery (see Fig.~\ref{fig:init_profile}c).
This region is influenced by complex particle dynamics arising from both upstream motion and contact interactions, making it more difficult to attribute observed behavior to local velocity features.
As a result, parameter reconstruction becomes relatively ill-posed and tends to degrade near the base of the domain. 
Nonetheless, the global approximation capability of the neural network parameterization maintains robust estimation even in this challenging region. 

These results demonstrate the effectiveness of JAX-MPM in learning spatially varying fields with neural-network-based representations.

\subsection{Inverse estimation of frictional coefficient}\label{sec:inverse_friction}

Frictional heterogeneity is prevalent in real-world geohazard scenarios and significantly affects flow mobility and deposition patterns. 
However, frictional properties are rarely measurable in situ,  making inverse modeling an essential tool for 
estimating such parameters from observable outcomes.

In this example, we demonstrate the ability of JAX-MPM to recover spatially varying Coulomb friction coefficients \( \mu_f \) along the bottom boundary of a 2D dam-break scenario using \textit{particle-trajectory data}. 

The simulation is initialized in a standard dam-break configuration within a domain of length \( L = 1.6\,\mathrm{m} \)  and height \( H = 0.4\,\mathrm{m} \), with an initial water column of \( L_0 = 0.5\,\mathrm{m} \) and \( H_0 = 0.3\,\mathrm{m} \). 
The bottom surface is divided into four equal-length segments, each assigned a ground-truth friction coefficient, i.e., 
$ (\mu_0, \mu_1, \mu_2 , \mu_3 ) = (0.0, 0.5, 0.1, 0.2)$, which are treated as trainable parameters during optimization. 
Since each segment interacts with the flow at different times, supervision is provided using particle trajectories sampled from \( |\mathcal{T}_t| = 1000 \) observation frames, uniformly distributed  over the full 10,000-step simulation (corresponding to 0.4\,s with a time step  of \( \Delta t = 4 \times 10^{-5} \,\mathrm{s} \)). Two  levels of observation sparsity are tested:
\( |\mathcal{T}_x| = 100 \) and \( |\mathcal{T}_x| = 500 \) randomly selected particles.

Frictional contact is modeled using the Coulomb friction law,
which modifies the tangential velocity at contact nodes based on particle–wall interaction. The corrected grid velocity \( \tilde{\boldsymbol{v}}_i^{n+1} \)  at contact node \( i \) is given by~\cite{bardenhagen2001improved}:

\begin{equation}
\tilde{\boldsymbol{v}}_i^{n+1}  = \boldsymbol{v}_i^{n+1} - 
\left[\Delta \boldsymbol{v} \cdot \boldsymbol{n}_i \right] 
\left( \boldsymbol{n}_i + \mu' \, \boldsymbol{n}_i \times \boldsymbol{\omega} \right),
\label{eq:coulomb_friction}
\end{equation}
where:
\begin{equation}
\Delta \boldsymbol{v} := \boldsymbol{v}_i^{n+1} - \boldsymbol{v}^{\text{cm}}_i, 
\quad
\boldsymbol{\omega} = \frac{\Delta \boldsymbol{v} \times \boldsymbol{n}_i}{\|\Delta \boldsymbol{v} \times \boldsymbol{n}_i\|}, \quad
\mu' = \min \left[\mu, \frac{\|\Delta \boldsymbol{v} \times \boldsymbol{n}_i\|}{\Delta \boldsymbol{v} \cdot \boldsymbol{n}_i} \right]
\end{equation}
where $\boldsymbol{v}^{\text{cm}}_i$ is the center-of-mass velocity of the two bodies at contact node $i$, which reduces to zero as one of the bodies is a stationary rigid wall \cite{gao2022mpm}.  $\mu$ is the friction coefficient, \( \boldsymbol{v}_i^{n+1} \) is the intermediate grid velocity before contact correction, \( \boldsymbol{n}_i \) is the normal vector at the contact surface (e.g., \( [0, -1] \) for bottom wall contact),
$\boldsymbol{\omega}$ is the unit tangential vector, and $\mu'$ is the effective friction coefficient.

\begin{figure}
    \centering
    \includegraphics[width=1.0\linewidth]{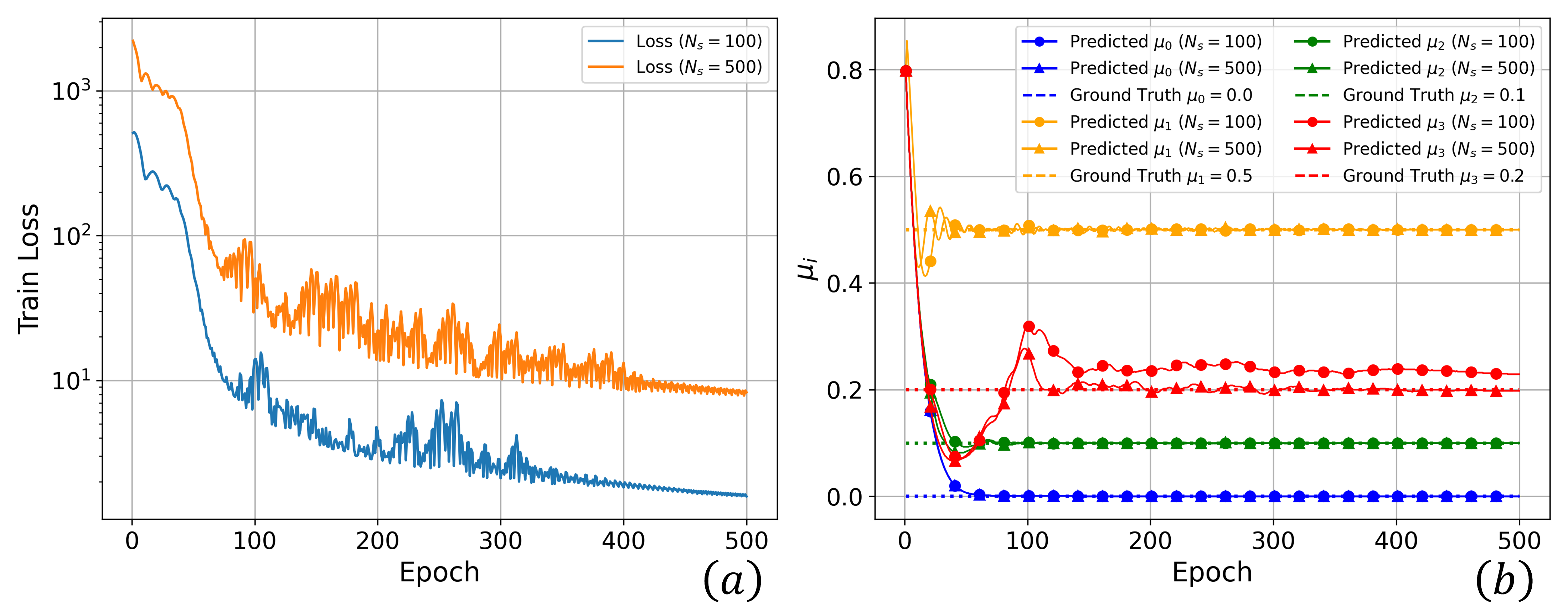}
    \caption{Inverse recovery of spatially varying Coulomb friction coefficients in a segmented dam-break domain. (a) Training loss over epochs, (b) Evolution of the predicted friction coefficients $\mu$ for each segment compared to the ground-truth values.}
    \label{fig:fric_multiple}
\end{figure}

Fig.~\ref{fig:fric_multiple}b shows the evolution of the four estimated friction coefficients under the two different observation densities throughout the optimization process. The first three upstream segments (\( \mu_0, \mu_1, \mu_2 \)) are accurately identified in both cases. 
The most notable discrepancy appears in the downstream segment \( \mu_3 \): with \( |\mathcal{T}_x| = 100 \), the estimated coefficient converges to 0.229 (14.5\% error), while with \( |\mathcal{T}_x| = 500 \), it converges to 0.198 (1.0\% error).
This highlights the role of observation density in improving parameter accuracy, particularly for downstream regions that interact less significantly with the  flow.

The convergence behavior reflects the cumulative nature of frictional effects. Upstream parameters, which influence a larger number of particle trajectories, generate stronger gradient signals and thus converge more rapidly. In contrast, downstream parameters interact with the flow later and more locally, resulting in weaker gradients, especially under sparse supervision, making them more challenging to accurately recover.

This example demonstrates that JAX-MPM can reliably recover spatially varying friction coefficients from partial trajectory data, even in contact-dominated systems with discontinuous boundary conditions.
Its differentiable Lagrangian formulation enables direct optimization of localized, non-smooth parameters from sparse observations. This capability is essential for real-world geomechanical applications, where frictional heterogeneity is common and observational data are often limited.

%% file: appendix.tex
\appendix
\section{Stress update for Drucker–Prager elastoplasticity}\label{appendix:dp}
This appendix details the Drucker–Prager constitutive model and its rate-independent return mapping update with Jaumann objective rate. The model parameters are defined as
\[
q_\varphi = \frac{6 \sin \varphi}{\sqrt{3} (3 + \sin \varphi)}, \quad
k_\varphi = \frac{6 c \cos \varphi}{\sqrt{3} (3 + \sin \varphi)}, \quad
q_\psi = \frac{6 \sin \psi}{\sqrt{3} (3 + \sin \psi)},
\]
where \( \varphi \) and \( \psi \) are the internal friction and dilation angles, respectively, and \( c \) is the cohesion. The model also uses the elastic shear and bulk moduli \( G \) and \( K \).

We adopt a rate-independent return-mapping algorithm that incorporates Drucker–Prager shear yielding and a tensile failure cutoff, with objectivity ensured via the Jaumann stress rate. The procedure involves the following steps \cite{huang2015large}:

\begin{enumerate}[(1)]
    \item \textit{Elastic trial stress.} For clarity, we reformulate the stress tensor in Eq. \eqref{eq:dp_stress} and other quantities using index notation. The trial stress at $t^{n+1}$ is evaluated by combining the rotated stress and the deformation increment:
    \begin{equation}
        {}^*\sigma_{ij}^{n+1} = \sigma_{ij}^{R n} + C_{ijkl} \Delta d_{kl},
    \end{equation}
    where $C_{ijkl}$ is the tangential stiffness tensor of the material, $\sigma_{ij}^{R n}$ is the Jaumann-rotated stress from the previous step:
    \begin{equation}
        \sigma_{ij}^{R n} = \sigma_{ij}^{n} + \sigma_{ik}^{n} \Delta \omega_{jk} + \sigma_{jk}^{n} \Delta \omega_{ik}.
    \end{equation}
    \item \textit{Trial invariants.} The trial spherical (mean) and effective shear stresses of the trial stress are given by:
    \begin{align}
        {}^*\sigma_m^{n+1} &= \frac{1}{3} {}^*\sigma_{kk}^{n+1}, \\
        {}^*\tau^{n+1} &= \sqrt{ \frac{1}{2} {}^*s_{ij}^{n+1} {}^*s_{ij}^{n+1} },
    \end{align}
    where ${}^*s_{ij}^{n + 1} ={}^*\sigma _{ij}^{n + 1} - {}^*\sigma _m^{n + 1}$ is the deviatoric components of trial stress.
    \item \textit{Yield conditions.} Plastic yielding is governed by the following three functions:
\begin{align}
    \text{Shear failure yield function:} \quad 
    f^s &= {}^*\tau^{n+1} - k_\varphi + q_\varphi {}^*\sigma_m^{n+1} \\
    \text{Tension failure yield function:} \quad 
    f^t &= {}^*\sigma_m^{n+1} - \sigma^t \\
    \text{Auxiliary transition function:} \quad 
    h &= {}^*\tau^{n+1} - \tau^P - \alpha^P ({}^*\sigma_m^{n+1} - \sigma^t)
\end{align}
where the auxiliary function \( h \) \cite{zhang2016material} defines a transition boundary between shear and tensile failure domains, with constants \( \tau^P \) and \( \alpha^P \) given by:
    \begin{equation}
        \tau^P = k_\varphi - q_\varphi \sigma^t\quad \text{and} \quad\alpha^P = \sqrt{1 + q_\varphi^2} - q_\varphi.
    \end{equation}
Based on the evaluated yield functions, the trial stress state is classified into three zones:

\begin{enumerate}
    \item Zone I (Elastic region):
    \begin{equation}
    f^s({}^*\sigma_m^{n+1}, {}^*\tau^{n+1}) \leq 0 \quad \text{and} \quad f^t({}^*\sigma_m^{n+1}) < 0
    \end{equation}
    No yielding occurs, and the trial stress is accepted without correction.

    \item Zone II (Shear failure region): 
    \begin{equation}
    \text{Either} \quad
    \begin{cases}
        f^s({}^*\sigma_m^{n+1}, {}^*\tau^{n+1}) > 0 \quad \text{and} \quad f^t({}^*\sigma_m^{n+1}) < 0 \\
        h({}^*\sigma_m^{n+1}, {}^*\tau^{n+1}) > 0 \quad \text{and} \quad f^t({}^*\sigma_m^{n+1}) \geq 0
    \end{cases}
    \end{equation}

    \item Zone III (Tensile failure region):
    \begin{equation}
    h({}^*\sigma_m^{n+1}, {}^*\tau^{n+1}) \leq 0 \quad \text{and} \quad f^t({}^*\sigma_m^{n+1}) \geq 0
    \end{equation}
\end{enumerate}
    
\item \textit{Shear failure correction (Zone II).} 

In Zone II, the trial stress violates the shear yield condition \( f^s > 0 \), indicating plastic shear deformation. The stress is projected back to the shear yield surface \( f^s = 0 \) using a radial return mapping algorithm under the assumption of incompressible plastic flow. The updated stress components are:
\begin{align}
    \sigma_m^{n+1} &= {}^*\sigma_m^{n+1} - K q_\psi \Delta \lambda^s, \\
    \tau^{n+1} &= k_\varphi - q_\varphi \sigma_m^{n+1}, \\
    \sigma_{ij}^{n+1} &= \frac{\tau^{n+1}}{{}^*\tau^{n+1}} \, {}^*s_{ij}^{n+1} + \sigma_m^{n+1} \delta_{ij},
\end{align}
and the equivalent plastic strain increment is:
\begin{equation}
    \Delta \varepsilon_{eq,p} = \Delta \lambda^s \sqrt{ \frac{1}{3} + \frac{2}{9} q_\psi^2 }.
\end{equation}

The plastic multiplier \( \Delta\lambda^s \) is determined by the consistency condition:
\( f^s=0 \)

\begin{equation}
    \Delta \lambda^s = \frac{f^s}{G + K q_\varphi q_\psi}.
\end{equation}

\item \textit{Tensile failure correction (Zone III).}  

In Zone III, where when the trial mean stress exceeds \( \sigma^t \) and the state lies outside the transition surface, the tensile failure occurs. A plastic correction is applied to cap the mean stress at \( \sigma^t \).
The updated stress and equivalent plastic strain are:
\begin{align}
    \sigma_{ij}^{n+1} &= {}^*\sigma_{ij}^{n+1} + \left( \sigma^t - {}^*\sigma_m^{n+1} \right) \delta_{ij}, \\
    \Delta \varepsilon_{eq,p} &= \frac{\sqrt{2}}{3} \Delta \lambda^t,
\end{align}

where the plastic multiplier \( \Delta \lambda^t \) is derived from the consistency condition \( f^t = 0 \):
\begin{equation}
    \Delta \lambda^t = \frac{{}^*\sigma_m^{n+1} - \sigma^t}{K}.
\end{equation}
\end{enumerate}

\section{Velocity transfer schemes}\label{appendix:pic}

JAX-MPM supports multiple grid-to-particle (G2P) transfer schemes, including PIC, FLIP, APIC, and TPIC. Among them, APIC~\cite{jiang2015affine} and TPIC~\cite{nakamura2023taylor} retain the same velocity update formulation as in PIC for computing the particle velocity \( \boldsymbol{v}_p^n \), but additionally incorporate particle-level information such as affine velocity fields or velocity gradients to reduce numerical dissipation and enhance stability.
These additional particle quantities (such as the affine matrix \( \boldsymbol{B}_p^n \) in APIC or the velocity gradient \( \nabla \boldsymbol{v}_p^n \) in TPIC) are evaluated during the G2P step and then used in the subsequent P2G step to compute the nodal velocity field.

In APIC, the nodal velocity at time step \( n \) is given by:
\begin{equation}
\boldsymbol{v}_i^n = \frac{1}{m_i^n} \sum_p \phi_{ip}^n m_p \left( \boldsymbol{v}_p^n + \boldsymbol{B}_p^n \cdot \left( \boldsymbol{D}_p^n \right)^{-1} \cdot (\boldsymbol{x}_i^n - \boldsymbol{x}_p^n) \right)
\end{equation}
where \( \boldsymbol{B}_p^n = \sum_i \phi_{ip}^{n-1} \boldsymbol{v}_i^{n} \otimes (\boldsymbol{x}_i^{n-1} - \boldsymbol{x}_p^{n-1}) \), and \( \boldsymbol{D}_p^n = \sum_i \phi_{ip}^n (\boldsymbol{x}_i^n - \boldsymbol{x}_p^n) \otimes (\boldsymbol{x}_i^n - \boldsymbol{x}_p^n) \) is the moment matrix.

In TPIC, the affine contribution is replaced by the particle velocity gradient \( \nabla \boldsymbol{v}_p^n \), which is derived from prior particle and grid states. Thus, the nodal velocity is computed as:
\begin{equation}
\boldsymbol{v}_i^n = \frac{1}{m_i^n} \sum_p \phi_{ip}^n m_p \left( \boldsymbol{v}_p^n + \nabla \boldsymbol{v}_p^n \cdot (\boldsymbol{x}_i^n - \boldsymbol{x}_p^n) \right)
\end{equation}